\pdfoutput=1

\documentclass[11pt]{article}

\usepackage[final]{acl}

\usepackage{times}
\usepackage{latexsym}

\usepackage{makecell}
\usepackage{pifont}
\usepackage{amsfonts}
\usepackage{amsmath}
\usepackage{subcaption}
\usepackage{multirow}
\usepackage{colortbl}
\usepackage{booktabs}
\usepackage{adjustbox}
\usepackage{caption}
\usepackage{tcolorbox}
\usepackage{placeins}

\usepackage[T1]{fontenc}

\usepackage[utf8]{inputenc}

\usepackage{microtype}

\usepackage{inconsolata}

\usepackage{graphicx}

\title{Beyond Easy Wins: A Text Hardness-Aware Benchmark for LLM-generated Text Detection}

\author{Navid Ayoobi \\
  University of Houston\\
  Houston, TX, USA \\
  \small{\texttt{nayoobi@cougarnet.uh.edu}}\\\And
  Sadat Shahriar \\
  University of Houston\\
  Houston, TX, USA \\
  \small{\texttt{sshahria@cougarnet.uh.edu}} \\\And
    Arjun Mukherjee \\
  University of Houston\\
  Houston, TX, USA \\
  \small{\texttt{arjun@cs.uh.edu}} \\
  }

\begin{document}
\maketitle
\begin{abstract}
We present a novel evaluation paradigm for AI text detectors that prioritizes real-world and equitable assessment. Current approaches predominantly report conventional metrics like AUROC, overlooking that even modest false positive rates constitute a critical impediment to practical deployment of detection systems. Furthermore, real-world deployment necessitates predetermined threshold configuration, making detector stability (i.e. the maintenance of consistent performance across diverse domains and adversarial scenarios), a critical factor. These aspects have been largely ignored in previous research and benchmarks. Our benchmark, \textbf{SHIELD}, addresses these limitations by integrating both reliability and stability factors into a unified evaluation metric designed for practical assessment. Furthermore, we develop a post-hoc, model-agnostic humanification framework that modifies AI text to more closely resemble human authorship, incorporating a controllable hardness parameter. This hardness-aware approach effectively challenges current SOTA zero-shot detection methods in maintaining both reliability and stability. (Data and code: \url{https://github.com/navid-aub/SHIELD-Benchmark})
\end{abstract}

\section{Introduction}
The pervasiveness of large language models (LLMs) is largely attributed to their exceptional ability to process, comprehend, and generate text that closely resembles human composition.
Current deployment paradigms exhibit substantial heterogeneity, encompassing interactive dialogue systems, content summarization \cite{wang-etal-2023-element}, question answering \cite{kamalloo-etal-2023-evaluating}, and sentiment assessment \cite{hou-etal-2024-progressive}.
Yet, despite their beneficial applications, LLMs expose new potential avenues for malicious exploitation.
Such harmful practices include, but are not limited to, automated disinformation dissemination \cite{vykopal-etal-2024-disinformation}, academic plagiarism and cheating \cite{Cotton03032024,wahle-etal-2022-large}, and the fabrication of deceptive reviews \cite{10.1145/3580305.3599502}.
Beyond deliberate misuse scenarios, the automated identification and filtration of LLM-generated content from training corpora has become imperative for preserving the integrity of contemporary human-generated information in training datasets \cite{10.1162/coli_a_00549}.
This process facilitates the development of models with current knowledge and mitigating the risk of cascading hallucinations in LLMs \cite{rawte-etal-2023-troubling}.

The subtlety of distinguishing recurring patterns in LLM-generated text renders human classification efforts scarcely better than chance \cite{uchendu-etal-2021-turingbench-benchmark,clark-etal-2021-thats,dou-etal-2022-gpt}.
Consequently, research emphasis has shifted toward the development of automatic detection tools.
Current detectors encounter several critical shortcomings that compromise their robustness and reliability.
Most prominently, their inability to generalize to out-of-distribution cases leads to failures when analyzing texts generated by unseen models or characterized by unfamiliar stylistic nuances \cite{kuznetsov-etal-2024-robust,10651296}.
Furthermore, detector efficacy is significantly diminished through minimal perturbations, text length modifications, or the application of adversarial techniques \cite{zhou-etal-2024-humanizing,huang-etal-2024-ai} including paraphrasing \cite{NEURIPS2023_30e15e59}, stylistic transformation, and intentional insertion of errors \cite{dugan2024raid}.

In efforts to improve detector robustness, most existing studies predominantly report conventional metrics, such as accuracy \cite{kuznetsov-etal-2024-robust}, F1-score \cite{guo2024biscope}, and AUROC \cite{NEURIPS2024_1d35af80,su2023detectllm,10.5555/3618408.3619446,bao2023fast} when assessing performance under diverse adversarial attacks.
However, this evaluation paradigm manifests several critical limitations in \textbf{real-world assessment} of detectors.
Primarily, \textbf{even modest false positive rates (FPR) are fundamentally unacceptable in LLM text detection contexts.}
For instance, in academic integrity applications, where the objective is to ensure fairness by identifying instances of academic dishonesty, misclassification of legitimately authored student work introduces significant procedural inequity by penalizing students undeservedly.
Consequently, some researchers have transitioned toward reporting true positive rates (TPR) at fixed FPR (e.g. $1\%$) \cite{hans2024spotting,yang2023dna}. 
However, despite this shift, additional unresolved issues remain.
When deployed in real-world applications, detection systems require configuration with a predetermined threshold, independent of the generative provenance of examined text.
Within this practical framework,\textbf{ quantifying detector stability} through analysis of \textbf{threshold dynamics} across diverse adversarial conditions becomes critically important, \textbf{a dimension that previous studies have largely overlooked}.
Consequently, aforementioned metrics provide inadequate characterization of practical detector efficacy.

In this paper, we propose novel evaluation metrics that facilitate more equitable comparative assessment of detection methods by simultaneously accounting for \textbf{FPR impact} on performance and detector \textbf{stability variation} across diverse scenarios.
We integrate these multidimensional considerations into a unified metric that comprehensively characterizes both the \textbf{reliability and stability} of detection systems under real-world implementation conditions.
In addition, we present a post-hoc, model-agnostic framework designed to steer LLM-generated texts toward more human-like word distributions across calibrated difficulty gradients.
This humanification process spans multiple hardness levels and is implemented through three key strategies: \textbf{a) Random meaning-preserving mutation}, \textbf{b) AI-flagged word swap}, and \textbf{c) Recursive humanification loop}. 
These strategies specifically target vulnerabilities in contemporary zero-shot detection approaches \cite{10.5555/3618408.3619446,bao2023fast}, which predominantly operate by perturbing texts and measuring token statistical properties.
By progressively diminishing these detection signals while maintaining semantic coherence, our framework provides increasingly sophisticated evaluation scenarios that advance detector robustness assessment and illuminate the limitations of current detection approaches.
In essence, this paper evaluates state-of-the-art detection systems using our hardness-aware benchmark (incorporating both challenging samples and our fairness-oriented metrics), offering a broader and more real-world evaluation framework that pushes detection efforts ``\textbf{beyond easy wins}''!
The core contributions of this paper are the following:
\begin{itemize}
    \item We formulate a novel evaluation paradigm that integrates both detector performance and stability while specifically penalizing elevated FPRs, thus ensuring fair and rigorous comparative assessments.
    \item We develop a model-agnostic generation framework that produces LLM-generated texts with controlled difficulty gradients to systematically evaluate detectors' performance.
    \item We compiled the largest dataset to date, consisting of both human-written and LLM-generated texts prior to adversarial manipulation, see Table \ref{Tab:benchmarks}.
\end{itemize}

\section{Related work}
\subsection{LLM-generated text detection}
Detection of LLM-generated text falls into three principal categories: \textbf{watermarking} \cite{pmlr-v202-kirchenbauer23a,liu2024adaptive,Panaitescu-Liess_Che_An_Xu_Pathmanathan_Chakraborty_Zhu_Goldstein_Huang_2025}, \textbf{supervised techniques} \cite{guo2024biscope,guo2024detective,abassy-etal-2024-llm,yu-etal-2024-text}, and \textbf{zero-shot approaches} \cite{hans2024spotting,ma-wang-2024-zero,yang2023dna,bao2023fast}.
\textbf{Watermarking} embeds imperceptible signals during text generation.
These approaches
fail to protect unknowing third-party users and are susceptible to paraphrasing attacks \cite{pang2024no}.
\textbf{Supervised techniques} train classifiers atop encoder-based backbones like RoBERTa \cite{10.1007/978-3-030-84186-7_31} using annotated corpora.
These approaches manifest considerable performance deterioration when applied to out-of-distribution contexts.
\textbf{Zero-shot methods} operate without training requirements, exploiting LM generative mechanisms through statistical indicators including log-likelihood \cite{gehrmann-etal-2019-gltr}, perplexity \cite{hans2024spotting}, token rank \cite{gehrmann-etal-2019-gltr,su2023detectllm}, and entropy \cite{10.5555/3053718.3053722}. 
Many approaches require generating alternative text versions to detect statistical deviations \cite{10.5555/3618408.3619446,yang2023dna}, a computationally intensive process.
This issue is mitigated through reducing required revisions, and efficient sampling methods \cite{bao2023fast,su2023detectllm}.

\subsection{Benchmarks for LLM text detection}
The literature presents multiple benchmarks for evaluating LLM-generated text detection, each with varying characteristics in scale, diversity, and evaluation methodology \cite{uchendu-etal-2021-turingbench-benchmark,yu2025your,pudasaini-etal-2025-benchmarking}.
RAID \cite{dugan2024raid} systematically examines robustness across multiple decoding strategies.
MAGE \cite{li2024mage} extends evaluation capabilities across a broader spectrum of LLMs.
DetectRL \cite{wu2024detectrl} focuses on vulnerability assessment through implementation of adversarial attacks and perturbations. 
M4GT-Bench \cite{wang2024m4gt} contributes a multilingual evaluation framework, and HC3 \cite{guo2023close} compiles one of the largest ChatGPT-centric datasets available. 
Despite these significant contributions, current benchmarks commonly lack samples with structured difficulty gradients and principled metrics that ensure fairness in practical comparisons.
Our benchmark, \textbf{SHIELD}, represents the first benchmark to incorporate humanified samples with graduated hardness levels.
Furthermore, SHIELD pioneers a fairness-aware evaluation methodology, thus filling critical gaps in the current evaluation paradigm.
Table \ref{Tab:benchmarks} provides a comparative analysis of our proposed benchmark against existing benchmarks in English.
The comparison covers critical aspects including pre-attack dataset size, diversity of writing styles, utilization of multiple LLMs, structured hardness levels, and the presence of fairness-oriented evaluation metrics.
\begin{table}[t]
\caption{Comparative analysis of LLM-generated text detection benchmarks.}
    \centering
    \resizebox{\columnwidth}{!}{%
    \begin{tabular}{lcccccc}
         \makecell{Benchmark \\ name} & \makecell{Human \\ samples}& \makecell{LLM\\ samples} & \makecell{Num of\\Styles} & \makecell{Multiple\\ LLMs} & \makecell{Hardness\\Levels?} & \makecell{Fair \\Metric?}\\\hline
\makecell{BUST;\\\cite{cornelius-etal-2024-bust}}&3.2k&22k&3&\ding{51}&\ding{55}&\ding{55} \\          
\makecell{DetectRL;\\\cite{wu2024detectrl}}&11.2k&11.2k&4&\ding{51}&\ding{55}&\ding{55} \\ 
\makecell{ESPERANTO;\\\cite{ayoobi2024esperanto}}&36k&36k&4&\ding{51}&\ding{55}&\ding{55} \\ 
 \makecell{HC3;\\\cite{guo2023close}}&59k&27k&1&\ding{55}&\ding{55}&\ding{55} \\ 
 \makecell{MAGE ;\\\cite{li2024mage}}&154k&295k&7&\ding{51}&\ding{55}&\ding{55} \\ 
\makecell{M4GT-Bench;\\\cite{wang2024m4gt}}&65k&88k&6&\ding{51}&\ding{55}&\ding{55} \\     
 \makecell{MGTBench;\\\cite{he2024mgtbench}}&3k&18k&3&\ding{51}&\ding{55}&\ding{55} \\     
\makecell{RAID;\\\cite{dugan2024raid}}&15k&509k&8&\ding{51}&\ding{55}&\ding{55} \\ 
 \makecell{\textbf{SHIELD}; \\ \textbf{(Ours)}}&87.5k&612.5k&7&\ding{51}&\ding{51}& \ding{51}\\ 
         \hline
    \end{tabular}
    }
    
   \label{Tab:benchmarks}
\end{table}

\section{SHIELD benchmark: data creation, humanification, and metric design}
This section introduces the methodology underlying our benchmark \textbf{SHIELD}  (\textbf{\underline{S}}calable \textbf{\underline{H}}ardness-\textbf{\underline{I}}nformed \textbf{\underline{E}}valuation of \textbf{\underline{L}}LM \textbf{\underline{D}}etectors).

\subsection{Data creation}
SHIELD comprises seven diverse writing styles: \textbf{semi-formal} discourse from Medium posts, \textbf{journalistic} reporting from news sources, \textbf{evaluative} content from Amazon reviews, \textbf{question-answering} text from Reddit's ELI5, \textbf{scientific} writing from arXiv abstracts, \textbf{partisan-persuasion} reporting from pink slime, and \textbf{expository} documentation from Wikipedia.
Please refer to Appendix \ref{app:data} for additional data characteristics.
To obtain the LLM-generated counterparts, we deployed seven models: Llama3.2-1b,  Llama3.2-3b, Llama3.1-8b \cite{grattafiori2024llama}, Mistral-7b \cite{jiang2023mistral7b}, Qwen-7b \cite{bai2023qwen}, Gemma2-2b, and Gemma2-9b \cite{team2024gemma} for rephrasing of human-written texts. 
Additional specifications regarding models and prompting are detailed in Appendix \ref{app:models}.
To guarantee human authorship, the dataset comprises exclusively pre-2021 data, predating the emergence of LLMs.
The SHIELD dataset contains 87.5k human-written documents and 612.5k LLM-generated samples before the application of adversarial techniques or humanification processes.
Complete statistical details are presented in Appendix \ref{app:stat}.

\subsection{Hardness-aware humanification}
The core hypothesis of our approach is to replace words that strongly indicate LLM authorship with words indicative of human authorship.
Initially, we quantify each word's impact on authorship inference by the following scoring function: 
\begin{equation}
    M\!I_i = \sum_{x}P(x{,}w_i)log(\frac{P(x|w_i)}{P(x)})
\end{equation}
where $x$ represents authorship, and the mutual information (MI) quantifies the extent to which observing the word $w_i$ shifts our probabilistic belief regarding the text's authorship. This scoring is performed by the ranker module illustrated in Figure \ref{fig:abcd}(a).
Subsequently, we partition the vocabulary into two subsets: AI-associated $\mathbb{A}$, and human-associated $\mathbb{H}$ vocabularies based on their usage frequencies $f_i$.
This separation reflects whether a word predominantly contributes to the distribution of AI- or human-written texts. For humanification of text, we implement three strategies: \textbf{a) Random meaning-preserving mutation (RMM)}, \textbf{b) AI-flagged word swap (AWS)}, and \textbf{c) Recursive humanification loop (RHL)}.
Please see Appendix \ref{app:samples} for a sample text from each strategy.
\subsubsection{Random meaning-preserving mutation}
This approach simulates scenarios wherein malicious users substitute random words to circumvent detection or when $\mathbb{A}$ and $\mathbb{H}$ are inaccessible.
Figure \ref{fig:abcd}.b illustrates this process.
Let $\mathcal{D}=\{w_1,...,w_n\}$ be an AI-generated text consisting of a sequence of words.
We define $\mathcal{S}\subset\mathcal{D}$ as the set of non-stop-words, $\mathcal{S}=\{w_i\in\mathcal{D}|w_i\notin \text{StopWords}\}$.
Next, we randomly sample a subset $\mathcal{M}\subset\mathcal{S}$ such that $|\mathcal{M}|=\lfloor p.|S| \rfloor$, with $p$ representing the sampling ratio.
Then, we construct a masked version $\mathcal{D}^{mask}\!=\!\mathbf{Mask}(\mathcal{D},\mathcal{M})$ by replacing each word $w_i\in\mathcal{M}$ in $\mathcal{D}$ with the special token \texttt{<mask>}.
$\mathcal{D}^{mask}$ is fed into a masking language model $\mathbf{f_{MLM}}$ which outputs a ranked list of predictions $\mathbf{f_{MLM}^{(i)}}(\mathcal{D}^{mask})$ at each masked position $i$.
For each $i$, the first candidate $\hat{w}_i$ with highest rank is selected such that it differs from original word in $\mathcal{D}$, $\hat{w}_i\neq w_i$.
Finally, the edited text $\mathcal{D}^{edit}$ is produced by replacing each $w_i\in \mathcal{M}$ with the corresponding $\hat{w}_i$:
\begin{equation}\label{eq:replace}
    \mathcal{D}^{edit}\! =\! \mathbf{Replace}(\mathcal{D},\{(w_i,\hat{w}_i)\}_{w_i \in \mathcal{M}}) 
\end{equation}

\subsubsection{AI-flagged word swap}
\begin{figure}[t]
    \centering
        \includegraphics[width=\linewidth]{ 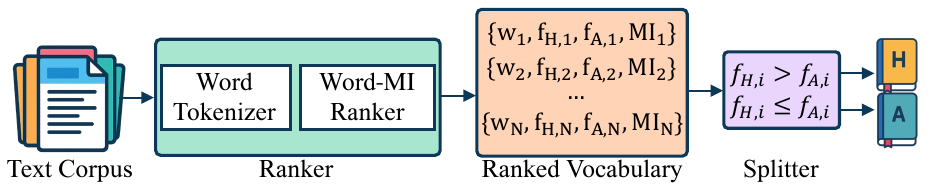}\caption*{(a)} 
    \hfill
        \includegraphics[width=\linewidth]{ 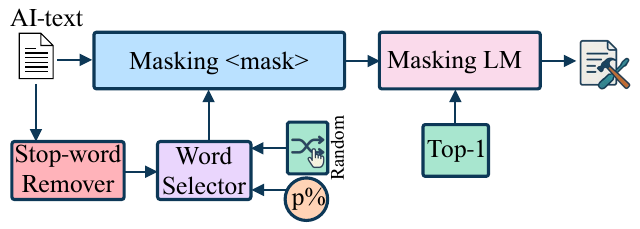}\caption*{(b)}
    \hfill
        \includegraphics[width=\linewidth]{ 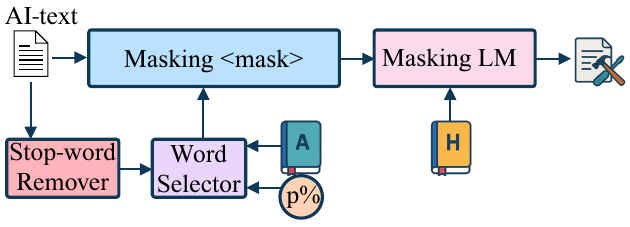}\caption*{(c)}
    \hfill
        \includegraphics[width=\linewidth]{ 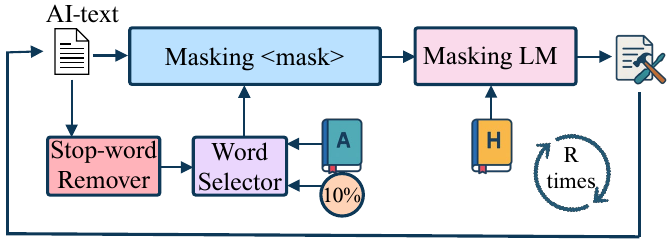}\caption*{(d)}
    \caption{Humanification strategies based on the ranked vocabularies $\mathbb{A}$ and $\mathbb{H}$ produced by the Ranker in (a). (b) Random meaning-preserving mutation (RMM), (c) AI-flagged word swap (AWS), (d) Recursive humanification loop (RHL).}
    \label{fig:abcd}
\end{figure}

The second strategy, depicted in Figure \ref{fig:abcd}(c), leverages $\mathbb{A}$ and $\mathbb{H}$ to substitute AI-indicative words with human-characteristic alternatives. 
Without loss of generality, let $\mathcal{S'}{=}\mathbf{Sort}(\mathcal{S},\text{MI}_{\mathbb{A}}){=}[w_{(1)},w_{(2)},...,w_{(|\mathcal{S}|)}]$ such that $\text{MI}_{\mathbb{A}}(w_{(1)}){\ge}\text{MI}_{\mathbb{A}}(w_{(2)}){\ge} ... {\ge}\text{MI}_{\mathbb{A}}(w_{(|\mathcal{S}|)})$.
Here, $\mathcal{S'}$ is the set $\mathcal{S}$ reordered in descending order based on MI scores with respect to the $\mathbb{A}$.
To construct set $\mathcal{M}$, we extract the top $p\%$ of words from $\mathcal{S'}$,
\begin{equation}
    \mathcal{M}{=}\{w_{(i)}\in \mathcal{S'}| 1\leq i \leq \lfloor p.|\mathcal{S}|\rfloor \}
\end{equation}
The parameter $p$ acts as a tunable knob that controls the hardness level of the humanified text. 
The words in $\mathcal{M}$ undergo masking in the initial text.
$\mathbf{f_{MLM}}$ then predicts candidate replacements for each masked position.
For each masked position $i$, we select the word with the highest score in $\mathbb{H}$,
\begin{equation}
    \hat{w}_i=\operatorname*{arg\,\,max}_{w \in \mathbf{f^{(i)}_{MLM}(\mathcal{D}^{mask})}}\!\!\!\text{MI}_{\mathbb{H}}(w)
\end{equation}
The edited text $\mathcal{D}^{edit}$ is derived similarly as Equation \ref{eq:replace}.

\subsubsection{Recursive humanification loop}
Figure \ref{fig:abcd}(d) shows the third strategy which extends second strategy (AWS) through implementation of a recursive refinement.
Let $\mathcal{D}^{(0)}$ be the original AI text.
We define the recursive editing process for $R$ rounds.
At each round $r\in \{1,2,...,R\}$, the set of non-stop words are formed,
\begin{equation}
    \mathcal{S}^{(r-1)}=\{w_i\in\mathcal{D}^{(r-1)}|w_i\notin \text{StopWords}\}
\end{equation}
Subsequently, words are ordered according to $\text{MI}_{\mathbb{A}}$ scores,
\begin{align}
    \mathcal{S'}^{(r-1)}&{=}\mathbf{Sort}(\mathcal{S}^{(r-1)},\text{MI}_{\mathbb{A}}) \notag \\
    &{=}[w_{(1)}^{(r-1)},w_{(2)}^{(r-1)},...,w^{(r-1)}_{(|\mathcal{S}|)}]
\end{align}
A fixed proportion $p{=}p_o$ (set at $10\%$) of words with maximal scores are selected for masking,
\begin{equation}
    \mathcal{M}^{(r)}{=}\{w^{(r-1)}_{(i)}\in \mathcal{S'}^{(r-1)}|1\leq i \leq \lfloor p_o.|\mathcal{S}|\rfloor \}
\end{equation}
$\mathbf{f_{MLM}}$ predicts the masked words in $D^{mask\,(r)}$, which is obtained by $\mathbf{Mask}(D^{(r-1)},\mathcal{M}^{(r)})$.
For each masked position $i$, we choose the word with the highest score in $\mathbb{H}$,
\begin{equation}
    \hat{w}^{(r)}_i=\operatorname*{arg\,\,max}_{w \in \mathbf{f^{(i)}_{MLM}(\mathcal{D}^{mask,\,(r)})}}\!\!\!\text{MI}_{\mathbb{H}}(w)
\end{equation}
Then the edited document for round $r$ is:
\begin{equation}\label{eq:replace2}
    \mathcal{D}^{(r)}\! =\! \mathbf{Replace}(\mathcal{D}^{(r-1)},\{(w_i,\hat{w}^{(r)}_i)\}_{w_i \in \mathcal{M}^{(r)}}) 
\end{equation}
The final humanified text is $\mathcal{D}^{edit}{=}\mathcal{D}^{(R)}$.
The parameter $R$ acts as a controllable knob to adjust the hardness level of the resulting text, with larger values yielding increasingly human-like phrasing.

\subsection{Fairness-oriented evaluation metric}
\subsubsection{Reliability and performance metric}
The AUROC metric can be interpreted as the probability that a randomly selected positive instance receives a higher score than a randomly selected negative one.
AUROC neglects consideration of practical operational threshold regions.
It uniformly weights the entire ROC curve, including high-FPR regions that are impractical for real-world deployment of AI-text detection systems.
It also fails to capture performance instability, that is, significant changes in TPR or FPR due to small threshold adjustments.
To more precisely evaluate the reliability of AI-text detection systems, we introduce weighted-AUROC (W-AUROC), defined as the expectation of TPR over a non-uniform probability distribution $p(t)$ across FPR,
\begin{equation}
    \text{W-AUROC}{=}\mathbb{E}_{t \sim p(t)}[\mathrm{TPR}(t)]
\end{equation}
where the weighting function is given by,
$p(t){=}\frac{1}{\mathcal{Z}}exp(-kt)$,
with decay parameter $k{>}0$ and normalization constant $\mathcal{Z}{=}\frac{1-exp(-k)}{k}$.
To determine the decay parameter $k$, we set the exponential weighting function $exp(-k.\mathrm{FPR})$ to decay to $50\%$ of its initial value at $\mathrm{FPR}{=}0.05$.
This decision is inspired by prior works in AI-text detection that routinely report TPR at a fixed FPR of less than $5\% $ as a key performance indicator, reflecting its status as a standard deployment-level operating point.
This constraint yields $k{=}20\,ln2$.
\subsubsection{Stability under FPR deviation (SFD)}
To assess stability across different scenarios, we compute the standard deviation of FPR at decision thresholds determined by Youden’s J statistic \cite{youden1950index} on ROC curve.
Our selection of FPR as the target variable stems from two considerations: 
1) the threshold determination in each detection system is intrinsically dependent on scoring function values with ranges varying between methods, 
2) this approach enables direct penalization of significant FPR fluctuations for stability assessment, as substantial FPR variability constitutes unacceptable performance in AI-text detection frameworks.
For each detection system and across $M$ evaluation scenarios (e.g., generative model, attack types, or writing styles), we extract the threshold $t^*_i$ that maximizes $J(t){=}\mathrm{TPR}(t)-\mathrm{FPR}(t)$  for each scenario $i$, and record the corresponding $\mathrm{FPR}^*_i{=}\mathrm{FPR}_i(t^*_i)$,
\begin{equation}
    t^*_i{=}\operatorname*{arg\,\,max}_{t}\,[\mathrm{TPR}_i(t)-\mathrm{FPR}_i(t)]
\end{equation}
The standard deviation of these optimal FPRs across all $M$ scenarios is calculated and denoted as $\sigma_{\mathrm{FPR}}$.
Finally, we define the stability metric as,
\begin{equation}
    \mathrm{SFD}{=}exp(-\lambda.\sigma_{\mathrm{FPR}})
\end{equation}
where $\lambda{>}0$ is a tunable hyperparameter controlling the sensitivity to instability. Lower standard deviation in FPR results in higher stability scores, with perfect stability ($\sigma_{\mathrm{FPR}}{=}0$) yielding a maximum value of 1.
To determine a principled value for the decay parameter $\lambda$, we calibrate it so that a moderate but practically noticeable instability in FPRs corresponds to a mid-range stability score.
Therefore, we set it such that the stability score reduces to $0.5$ when the standard deviation $\sigma_{\mathrm{FPR}}$ reaches $0.1$.
This yields $\lambda{=}10\,ln2$.

We adopt a multiplicative formulation to combine W-AUROC (performance) and SFD (stability) into a single unified reliability-stability score (URSS),
\begin{equation}
    \mathrm{URSS}=(\frac{1}{M}\sum_{i=1}^{M}\text{W-AUROC}_i)\,.\,\mathrm{SFD}
\end{equation}
Multiplication enforces a \textbf{non-compensatory relationship}: a high W-AUROC cannot mask poor stability, and conversely, robust stability does not necessarily indicate high discrimination capability.
This reflects real-world deployment requirements, where even a highly accurate detector is unusable if unstable, and vice versa.
\begin{figure*}[t]
    \centering
    \setlength{\tabcolsep}{2pt}
    \renewcommand{\arraystretch}{1}
    \begin{tabular}{cccc}
        \small{Medium} & \small{News} & \small{arXiv} & \small{Pink slime}\\
        \includegraphics[width=0.2\textwidth]{ 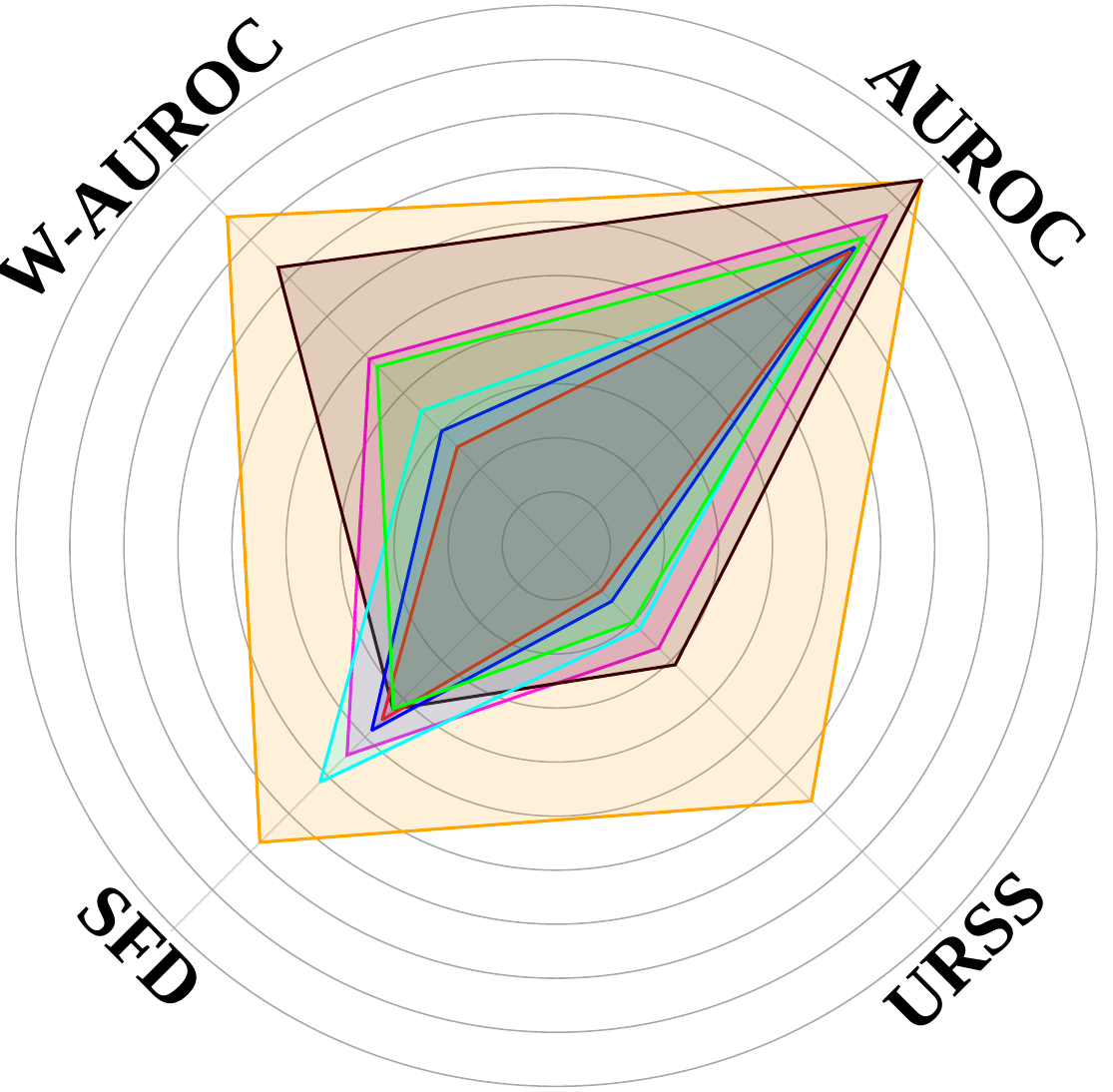} &
        \includegraphics[width=0.2\textwidth]{ 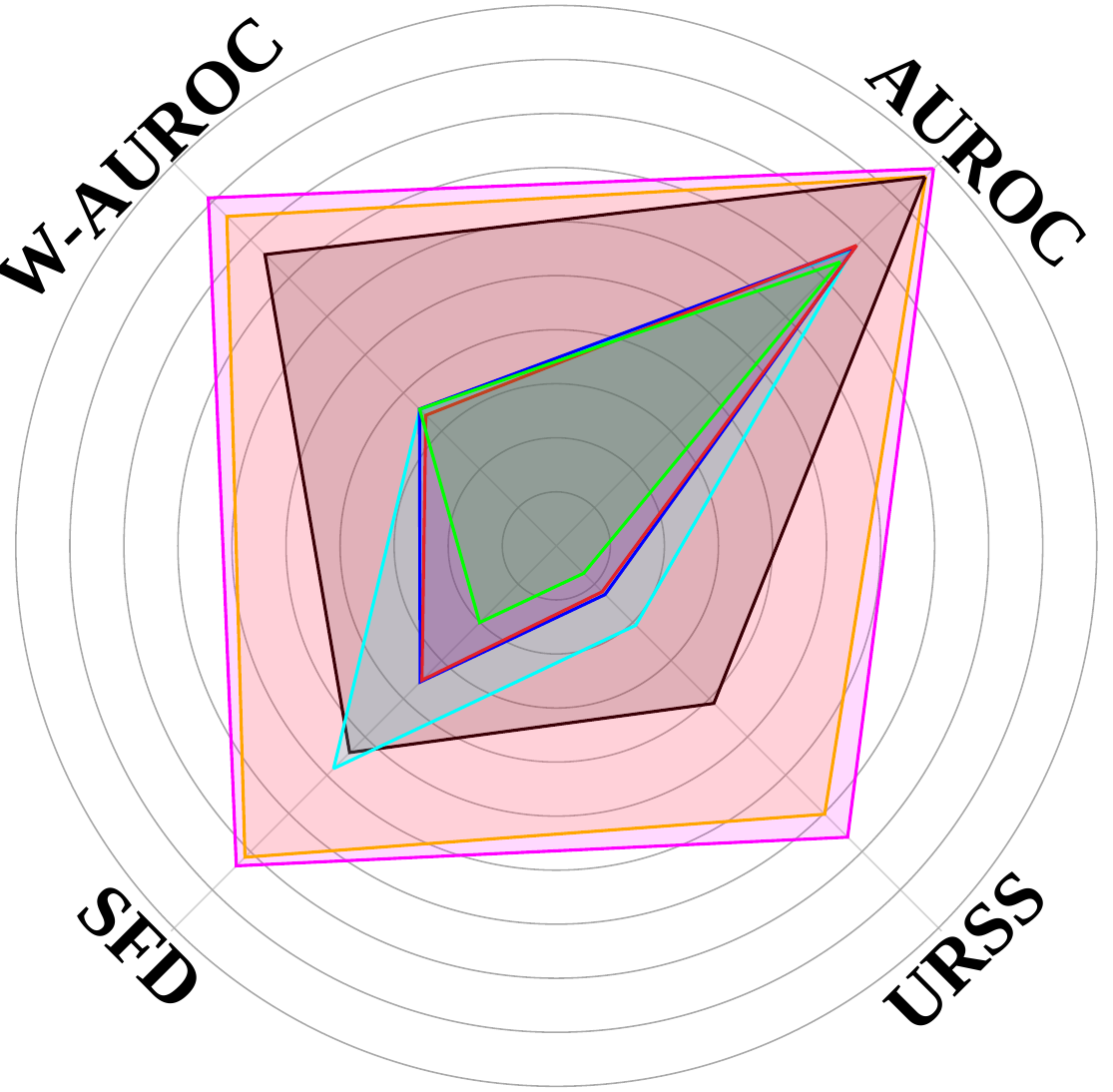} &
        \includegraphics[width=0.2\textwidth]{ 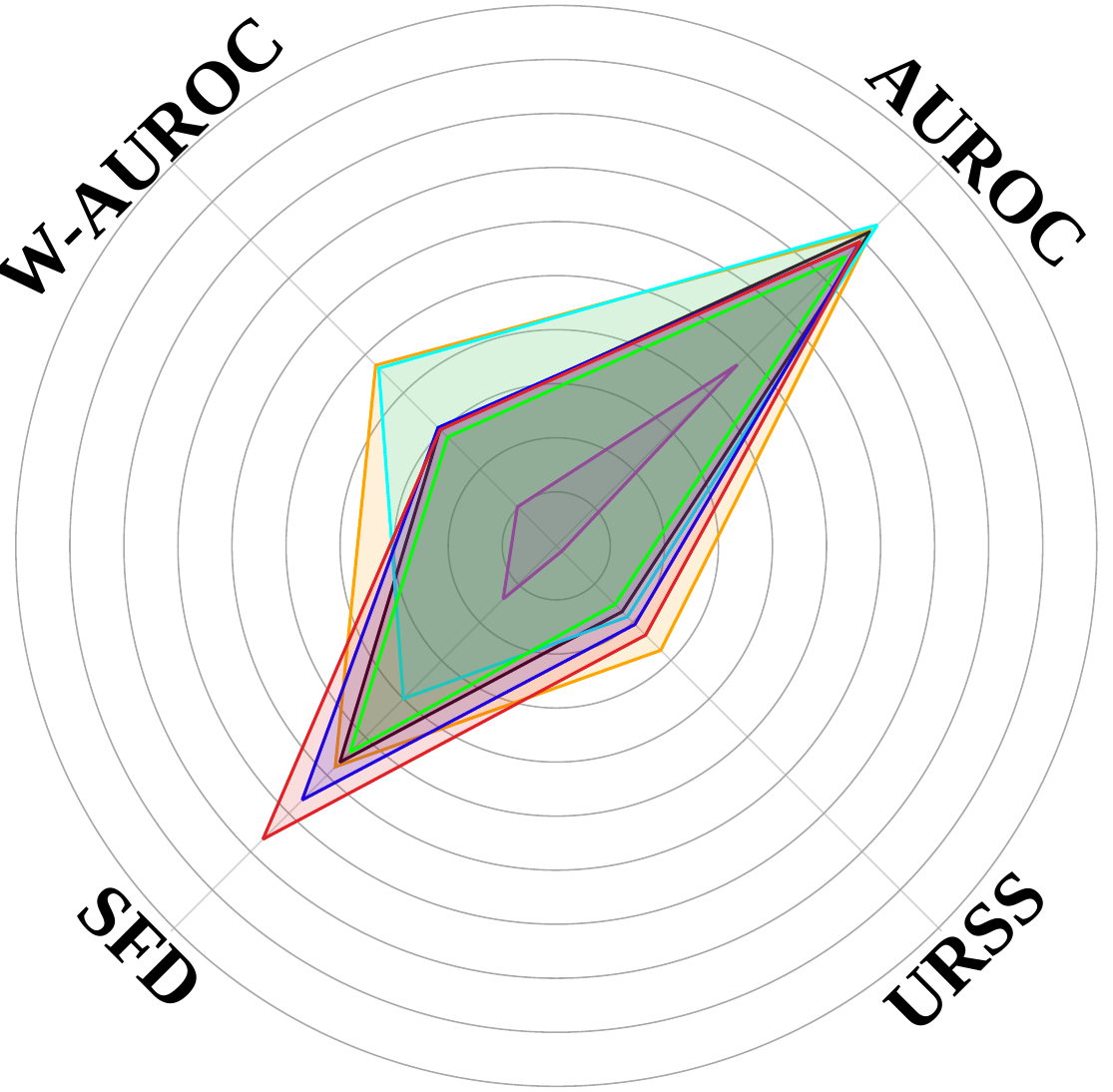} &
        \includegraphics[width=0.2\textwidth]{ 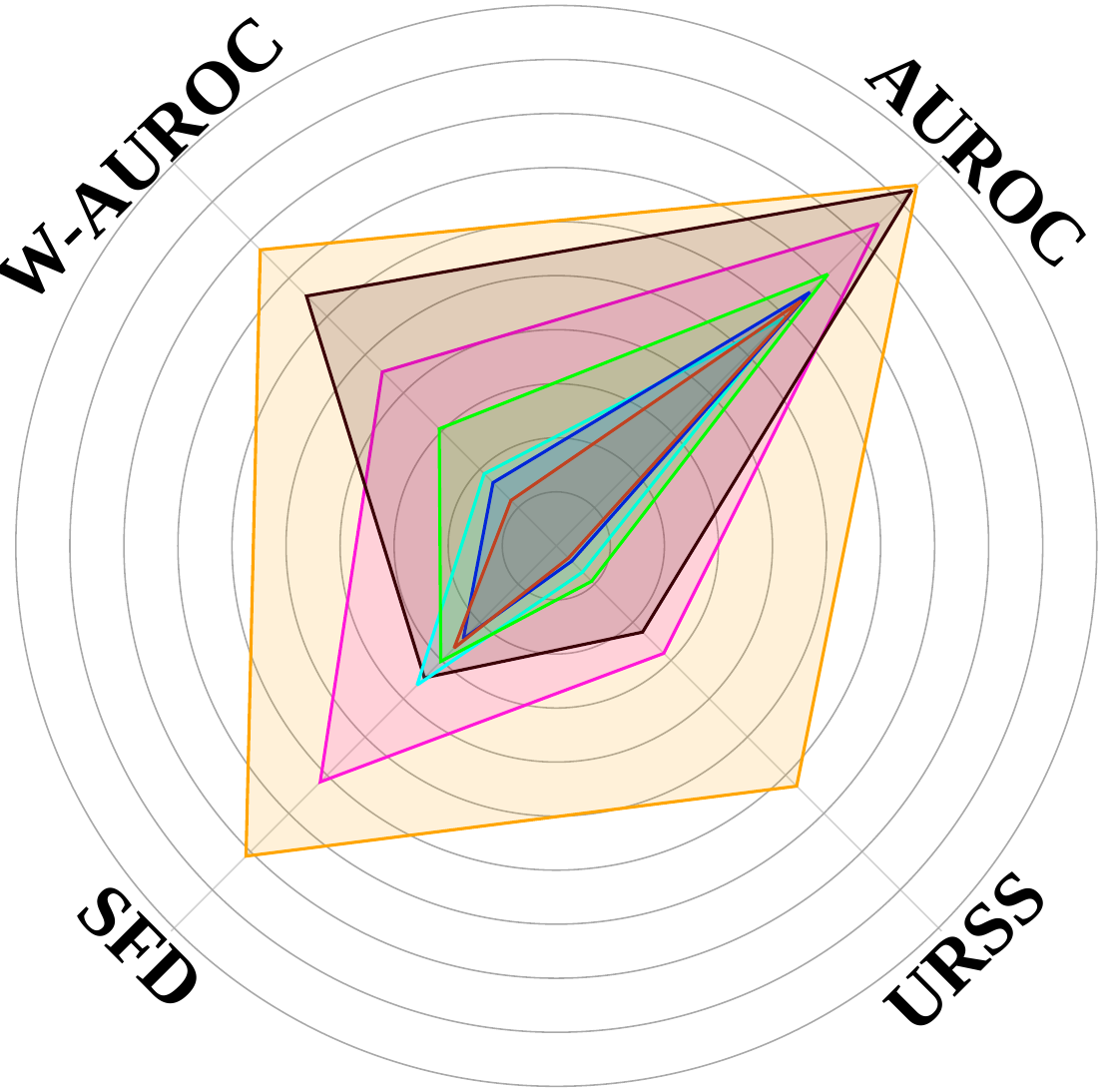}  \\
        \small{Reddit}& \small{Review} & \small{Wikipedia}\\
        \includegraphics[width=0.2\textwidth]{ 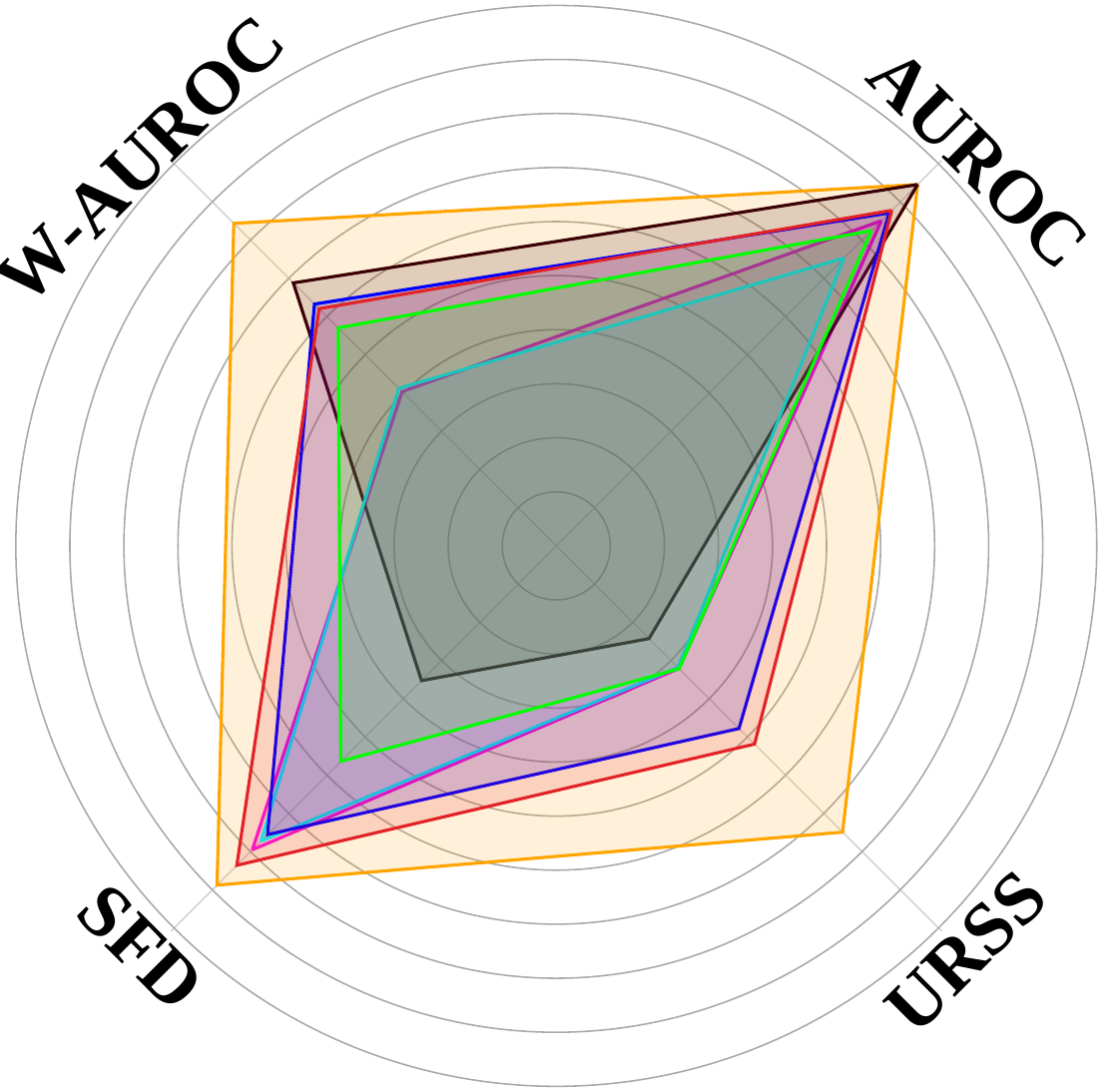} &
        \includegraphics[width=0.2\textwidth]{ 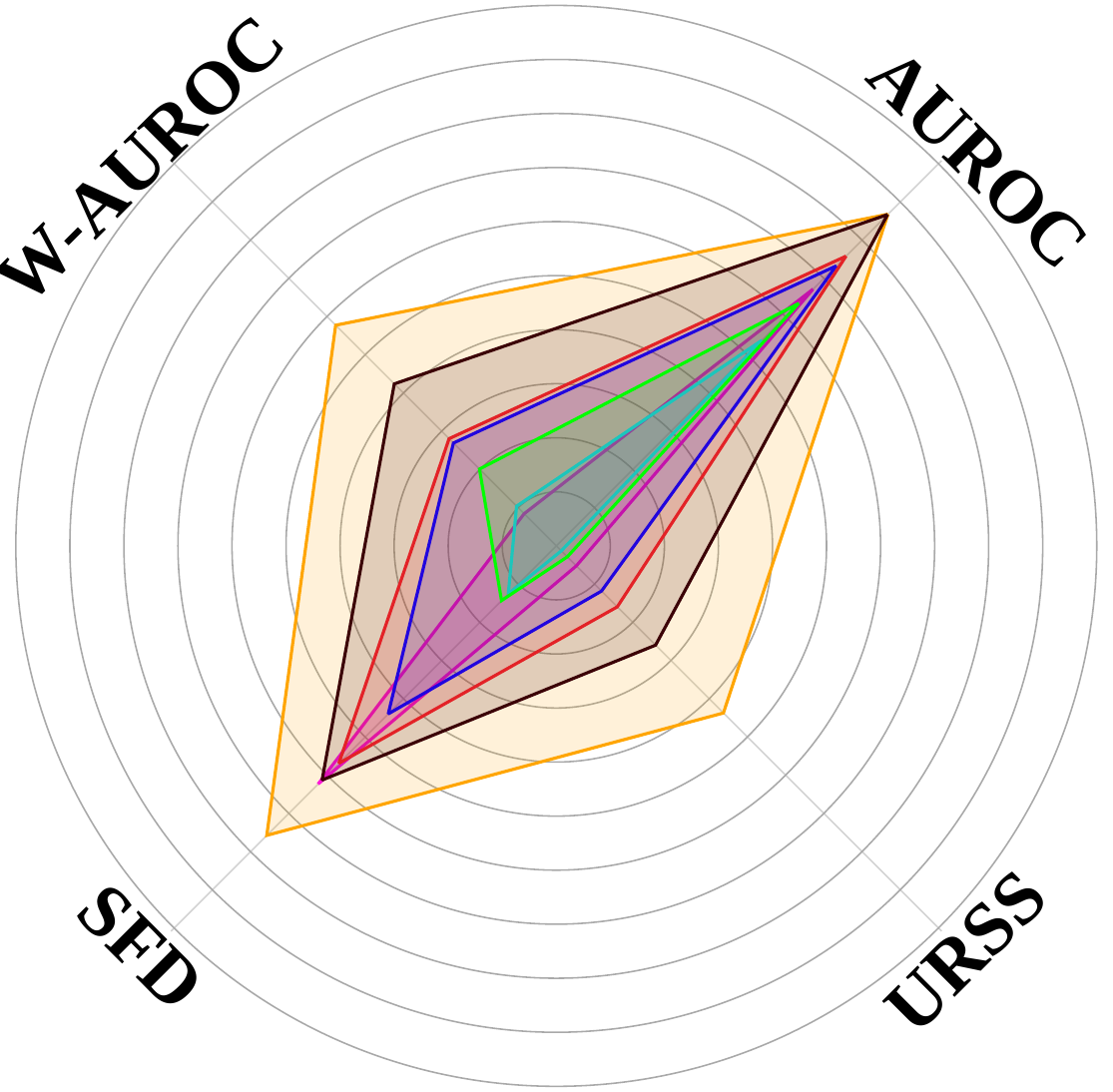} &
        \includegraphics[width=0.2\textwidth]{ 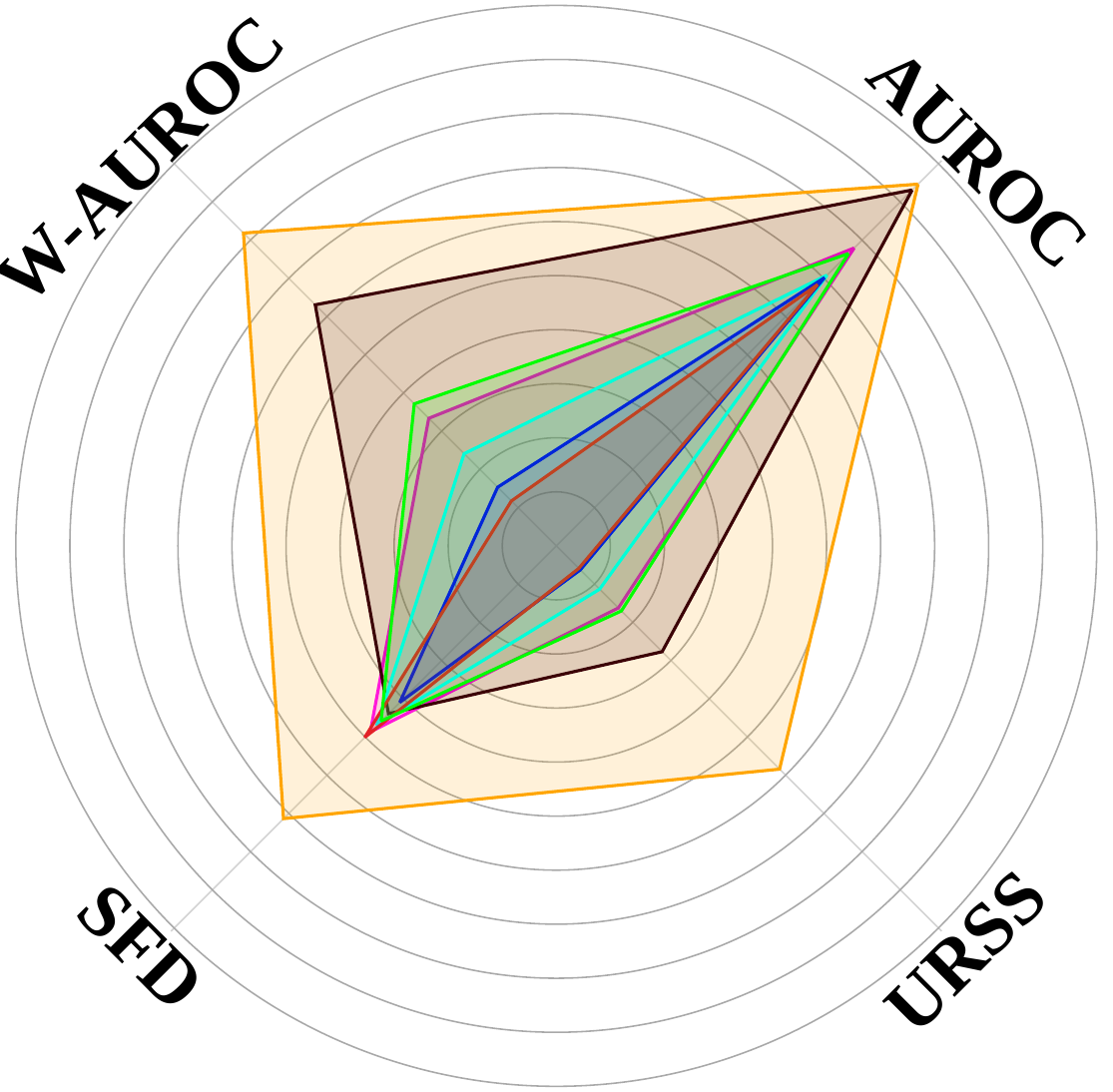} &
        \includegraphics[width=0.2\textwidth]{ 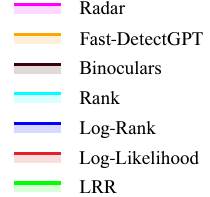} \\
   
    \end{tabular}

    \caption{Radar charts of comparing detectors in different writing styles across all generative models.}
    \label{fig:radarchart}
\end{figure*}
\begin{table*}[t]
\centering
\caption{Performance of detectors under paraphrasing (baseline), RMM, AWS, and RHL strategies.}
 \renewcommand{\arraystretch}{0.6}
\begin{adjustbox}{max width=\textwidth}
\begin{tabular}{l|cccc|cccc|cccc|cccc}
\toprule
\textbf{LLM $\rightarrow$} & \multicolumn{4}{c}{\textbf{Gemma2 9b}} & \multicolumn{4}{c}{\textbf{Llama3.1 8b}} & \multicolumn{4}{c}{\textbf{Mistral 7b}} & \multicolumn{4}{c}{\textbf{Qwen 7b}} \\
\textbf{Detector $\downarrow$ Metric ($\mathbf{\%}$) $\rightarrow$} & AUC & W-A & SFD & URSS & AUC & W-A & SFD & URSS & AUC & W-A & SFD & URSS & AUC & W-A & SFD & URSS \\
\midrule
\multicolumn{17}{c}{\cellcolor{yellow!15}\textbf{Paraphrase (baseline)}} \\
\midrule
Binoculars      &93.8& 69.9& 47.0& 32.9&95.4& 74.0& 55.2& 40.9&90.6& 59.4& 46.2& 27.4&85.5& 35.2& 42.5& 14.9\\
Fast-DetectGPT  & 93.1& 74.7& 60.7& 45.3&94.6& 80.3& 60.5& 48.6&89.2& 62.1& 62.1& 38.6&91.7& 76.2& 81.2& 61.8\\
Log-Likelihood &77.6& 28.7& 41.2& 11.8&82.2& 36.1& 41.8& 15.1&64.8& 16.6& 22.8& 3.8&66.0& 18.0& 42.6& 7.7\\
Log-Rank        & 77.6& 30.1& 47.3& 14.3&82.5& 38.6& 44.5& 17.2&63.3& 15.7& 20.8& 3.3&65.6& 18.1& 37.6& 6.8\\
LRR             & 76.2& 34.0& 58.4& 19.9&81.5& 45.1& 55.5& 25.0&58.5& 13.5& 16.9& 2.3&62.5& 17.8& 27.2& 4.8\\
Radar            & 76.7& 34.2& 17.6& 6.0&79.8& 42.3& 14.8& 6.3&71.0& 25.5& 15.9& 4.0&83.7& 50.7& 21.1& 10.7\\
Rank            &75.9& 35.6& 49.3& 17.5&76.3& 36.8& 57.1& 21.0&62.0& 16.9& 16.3& 2.8&59.3& 18.1& 16.0& 2.9\\
\midrule
\multicolumn{17}{c}{\cellcolor{yellow!15}\textbf{Random meaning-preserving mutation (RMM)}} \\
\midrule
Binoculars      &77.9& 35.4& 35.1& 12.4&83.6& 44.7& 43.8& 19.6&73.9& 28.9& 53.0& 15.3&77.7& 30.4& 46.2& 14.1\\
Fast-DetectGPT  & 77.0& 37.8& 35.8& 13.5&82.2& 47.8& 46.3& 22.1&71.7& 27.2& 45.8& 12.5&83.5& 50.4& 52.8& 26.6\\
Log-Likelihood &34.4& 4.7& 5.9& 0.3&41.2& 8.4& 6.2& 0.5&26.1& 2.2& 4.0& 0.1&31.3& 4.8& 8.1& 0.4 \\
Log-Rank        &36.9& 5.4& 7.3& 0.4&44.2& 9.6& 6.9& 0.7&27.6& 2.2& 4.5& 0.1&32.9& 4.8& 7.6& 0.4 \\
LRR             & 50.4& 9.6& 11.8& 1.1&58.5& 16.4& 10.3& 1.7&39.4& 3.4& 8.9& 0.3&43.1& 5.4& 5.7& 0.3 \\
Radar            & 89.0& 52.4& 34.0& 17.8&88.1& 52.3& 27.8& 14.5&83.7& 41.9& 24.8& 10.4&89.4& 60.7& 29.9& 18.2 \\
Rank            &48.4& 7.0& 9.5& 0.7&50.1& 8.4& 9.2& 0.8&39.7& 2.8& 6.7& 0.2&38.8& 3.3& 6.0& 0.2\\
\midrule
\multicolumn{17}{c}{\cellcolor{yellow!10}\textbf{AI-flagged word swap (AWS)}} \\
\midrule
Binoculars      &81.6& 45.4& 54.4& 24.7&81.9& 46.1& 37.2& 17.2&78.6& 41.2& 57.4& 23.6&73.2& 31.3& 42.6& 13.3\\
Fast-DetectGPT  &83.1& 45.9& 31.0& 14.2&83.3& 49.6& 36.3& 18.0&79.5& 37.7& 37.8& 14.3&78.5& 42.5& 43.1& 18.3\\
Log-Likelihood & 21.5& 2.5& 4.3& 0.1&25.0& 4.7& 5.3& 0.3&16.1& 1.3& 3.7& 0.0&18.8& 2.7& 18.4& 0.5 \\
Log-Rank        &22.7& 2.7& 4.3& 0.1&26.4& 5.1& 5.3& 0.3&16.7& 1.2& 4.0& 0.0&18.8& 2.4& 13.6& 0.3 \\
LRR             & 32.6& 4.0& 7.8& 0.3&36.1& 6.8& 4.9& 0.3&24.7& 1.4& 4.3& 0.1&22.3& 1.7& 3.7& 0.1\\
Radar            &85.9& 47.6& 30.0& 14.3&82.3& 40.3& 26.1& 10.5&80.4& 39.0& 19.0& 7.4&86.0& 52.3& 26.5& 13.9\\
Rank            &46.1& 4.9& 10.3& 0.5&46.5& 5.5& 10.5& 0.6&39.5& 2.4& 11.9& 0.3&31.3& 2.0& 5.7& 0.1\\
\midrule
\multicolumn{17}{c}{\cellcolor{yellow!10}\textbf{Recursive humanification loop (RHL)}} \\
\midrule
Binoculars      &75.8& 32.2& 59.3& 19.1&78.0& 36.2& 40.0& 14.5&75.0& 32.3& 63.4& 20.5&72.6& 26.6& 44.9& 11.9\\
Fast-DetectGPT  & 76.5& 30.9& 50.2& 15.5&78.2& 37.3& 42.4& 15.8&74.8& 27.1& 50.2& 13.6&77.4& 36.7& 61.0& 22.4 \\
Log-Likelihood &17.5& 0.6& 3.5& 0.0&21.6& 1.8& 3.8& 0.1&12.4& 0.5& 3.3& 0.0&17.2& 1.9& 8.3& 0.2\\
Log-Rank        &20.2& 0.8& 3.7& 0.0&24.5& 2.4& 4.2& 0.1&14.4& 0.5& 3.2& 0.0&18.3& 1.8& 6.7& 0.1\\
LRR             &36.4& 3.3& 11.6& 0.4&40.6& 6.4& 5.8& 0.4&29.4& 1.1& 10.9& 0.1&27.6& 1.7& 3.6& 0.1\\
Radar            &85.4& 43.9& 34.5& 15.1&81.8& 37.9& 27.0& 10.2&79.9& 33.9& 27.1& 9.2&86.6& 51.2& 30.9& 15.8 \\
Rank            &45.1& 4.4& 8.8& 0.4&45.1& 4.5& 9.0& 0.4&39.0& 1.7& 7.2& 0.1&32.6& 1.9& 4.5& 0.1\\
\bottomrule
\end{tabular}
\end{adjustbox}
\label{Tab:strategieslarge}
\end{table*}
\section{Experiments and discussion}
In this section, we organize our experiments and discussion around three core questions: 1) \textbf{Why is AUROC insufficient} for evaluating detectors in real-world settings? 2) \textbf{How effective is our proposed humanification strategies} in degrading SOTA zero-shot detectors? and 3) How robust are these detectors \textbf{across different levels of humanification hardness}?
We evaluate six SOTA zero-shot detectors, Binoculars \cite{hans2024spotting}, Fast-DetectGPT \cite{bao2023fast}, LRR \cite{su2023detectllm}, Log-Likelihood, Log-Rank, and Rank \cite{gehrmann-etal-2019-gltr}, along with a supervised model, Radar \cite{NEURIPS2023_30e15e59}, to examine the comparative \textbf{vulnerability of zero-shot methods}.
Further information on the detectors used can be found in Appendix \ref{app:detectors}.

\subsection{Why is AUROC NOT enough?}\label{sec:why}
For each writing style, we compare detectors across all generative models using the original LLM-generated texts (without humanification).
The resulting radar charts are presented in Figure \ref{fig:radarchart}.
Please refer to Appendix \ref{app:smallmodel} for radar charts of different LLMs across all writing styles.
While traditional \textbf{AUROC can exaggerate the superiority of a detector}, our proposed framework reveals an illuminating truth.
AUROC may obscure cases where detectors perform equivalently in practice.
In contrast, URSS effectively exposes equivalences by examining performance parity within operationally low FPR regions and assessing stability.
For instance, in the Reddit chart, although Radar exhibits a substantially higher AUROC than Rank, both achieve identical URSS, highlighting their practical equivalence.
Conversely, there are cases where detectors achieve similar AUROC scores, yet one outperforms the other in low-FPR operational regions while also exhibiting greater threshold stability. 
\textbf{Such multidimensional superiority is entirely masked when relying solely on the conventional AUROC metric}.
Representative cases include Rank vs. Log-Rank and Log-Rank vs. Log-Likelihood in the Medium chart, and Binoculars vs. Fast-DetectGPT across News, Wikipedia, Medium, and Review charts.
Despite significant disparities in either W-AUROC or SFD, identical AUROC across detectors can cause a misleading impression of equivalence. 
This potentially results in \textbf{suboptimal choices for real-world deployment}.
Such patterns are seen in Fast-DetectGPT vs. Rank and Log-Likelihood vs. Log-Rank on arXiv, Log-Rank vs. Log-Likelihood in Reddit, and Rank vs. Log-Likelihood in Wikipedia.
URSS suggests that \textbf{meaningful performance equivalency} between detectors can only be established through comprehensive evaluation that simultaneously considers both operational region sensitivity and cross-scenario stability.
For example, in the Wikipedia chart, URSS observes that Log-Likelihood has slightly lower W-AUROC but higher SFD than Log-Rank, and assigns them equal scores, reflecting fairness when each method excels in one dimension and the trade-off is not substantial.

A significant observation from our experiments is that methods exhibiting severe deficiencies in any critical performance dimension are heavily penalized, acknowledging that such limitations undermine real-world deployment potential.
This evaluation principle is exemplified by Binoculars which, while achieving the highest AUROC and second-highest W-AUROC scores, was ultimately positioned last in the Reddit writing style analysis based on URSS metric due to its exceptionally poor stability.

\subsection{Effectiveness of humanification strategies}
Table \ref{Tab:strategieslarge} shows the impact of different strategies on detector performance across all writing styles, using the largest LLM model from each family evaluated in this study.
Corresponding results for smaller models are available in Appendix \ref{app:smallmodel}.
To enhance readability, all metrics are reported as percentages. 
The baseline consists of original LLM-generated texts (specifically paraphrased texts without humanification).
Zero-shot detectors suffer marked average URSS degradation of $41\%$, $62\%$, $97\%$, $96\%$, $92\%$, and $95\%$ in \textbf{RMM}; $27\%$, $66\%$, $98\%$, $98\%$, $98\%$, and $95\%$ in \textbf{AWS}; and $38\%$, $65\%$, $99\%$, $99\%$, $97\%$, and $97\%$ in \textbf{RHL}, corresponding to Binoculars, Fast-DetectGPT, Log-Likelihood, Log-Rank, LRR, and Rank, respectively.
This demonstrates that without devising robustness enhancements, \textbf{zero-shot detectors fail to withstand word-level humanification}, resulting in severe performance loss.
Interestingly, even though AWS and RHL introduce more complex humanification than RMM, our experiments reveal that multiple detectors demonstrate similarly compromised URSS performance under the simpler RMM. 
For instance, Log-Likelihood’s URSS drops to $0.1$ under RMM for Mistral 7b, nearly equivalent to its $0.0$ under AWS and RHL.
This observation indicates that zero-shot detectors depend on fragile token-level statistical signatures that collapse under even modest perturbations. 
This vulnerability becomes particularly concerning considering that RMM more closely approximates natural user editing behaviors, rendering these detection systems unreliable in practical deployment scenarios.
This limitation exacerbates the previously documented challenge of generative model dependency \cite{wu2024detectrl}, as further evidenced in Table \ref{Tab:strategieslarge}, which demonstrates significant variation in URSS metrics across different LLMs when subjected to identical strategies.

It is noteworthy that Radar exhibits a reversed trend: across all humanification strategies, word replacements improve its performance in both W-AUROC and SFD, leading to higher URSS.
This may be attributed to Radar’s design, which emphasizes robustness to paraphrasing.
We leave further investigation into whether this behavior generalizes to other supervised methods for future work.

\subsection{Robustness under hardness levels}
\begin{figure*}[t]
    \centering
    \begin{tabular}{ccc}
    \small{RMM, AUROC} & \small{AWS, AUROC} & \small{RHL, AUROC} \\
    \includegraphics[width=0.3\textwidth]{ 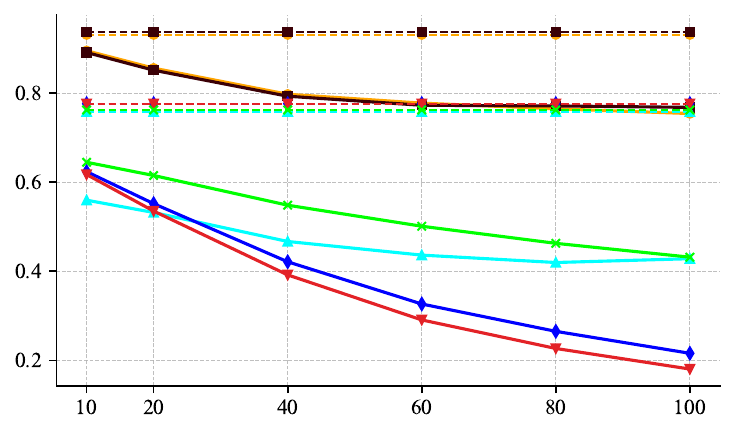} &
    \includegraphics[width=0.3\textwidth]{ 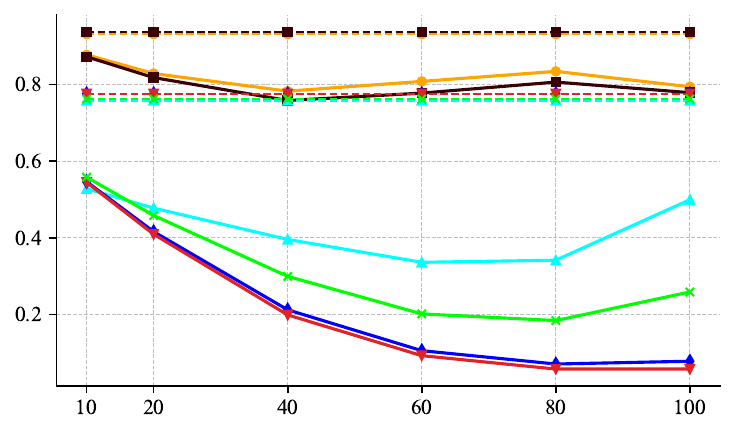} &
    \includegraphics[width=0.3\textwidth]{ 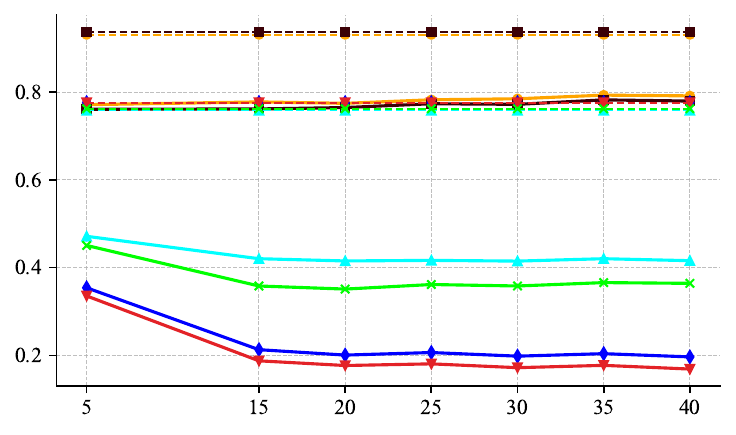}\\
    \small{RMM, W-AUROC} & \small{AWS, W-AUROC}  & \small{RHL, W-AUROC}\\
    \includegraphics[width=0.3\textwidth]{ 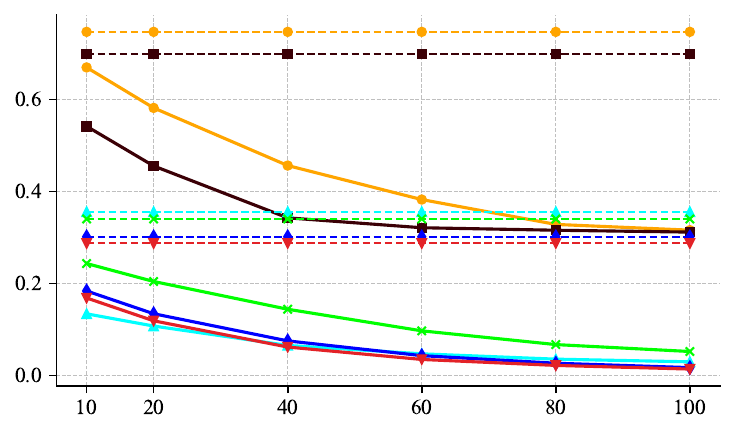} &
     \includegraphics[width=0.3\textwidth]{ 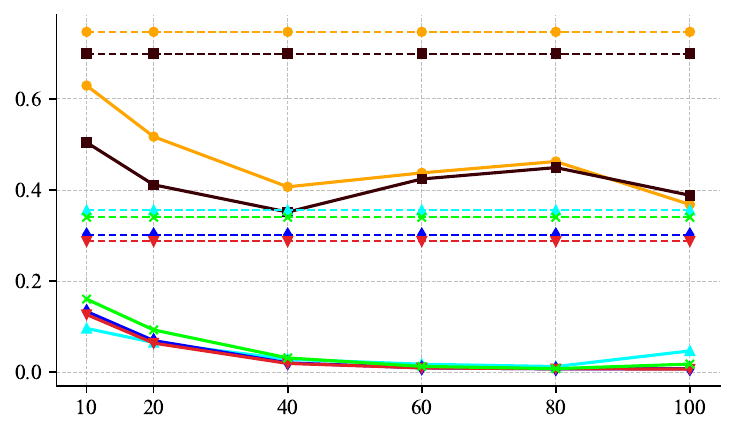} &
      \includegraphics[width=0.3\textwidth]{ 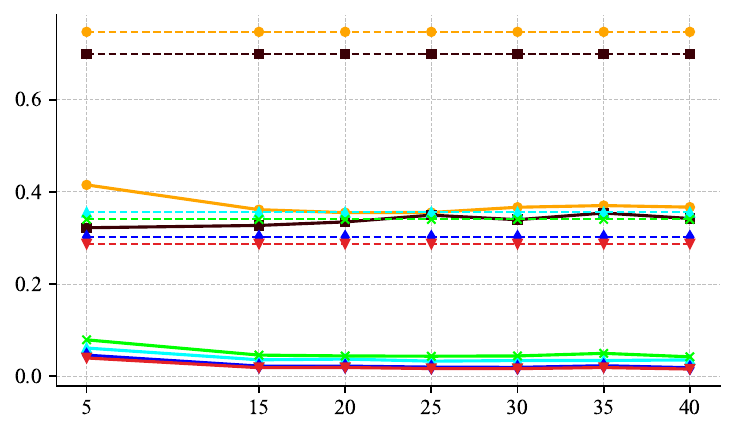} \\
    \small{RMM, SFD} & \small{AWS, SFD} & \small{AWS, URSS} \\
    \includegraphics[width=0.3\textwidth]{ 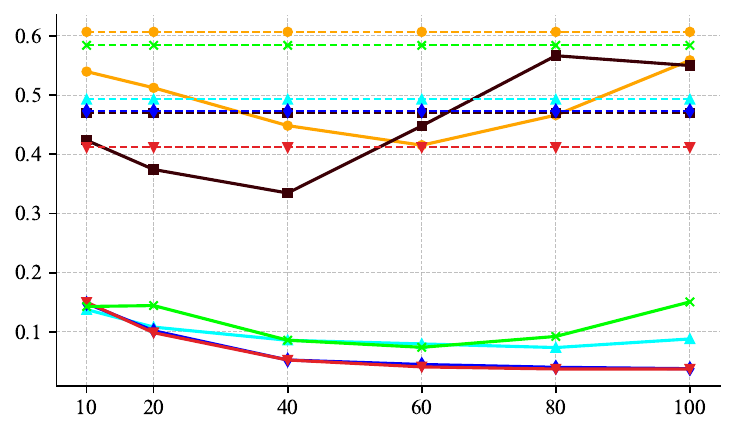} &
    \includegraphics[width=0.3\textwidth]{ 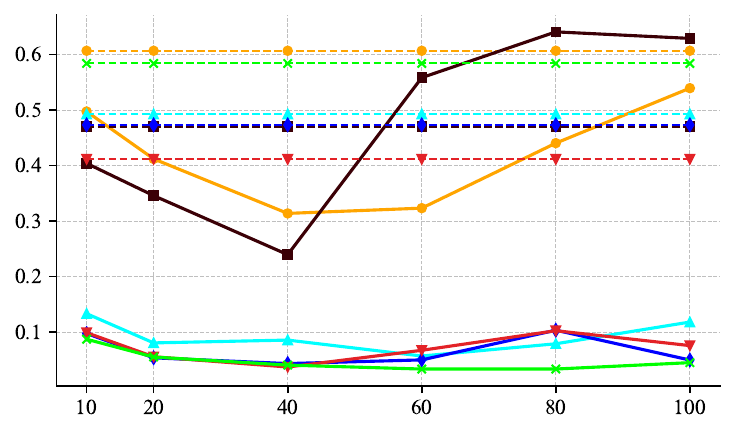} &
    \includegraphics[width=0.3\textwidth]{ 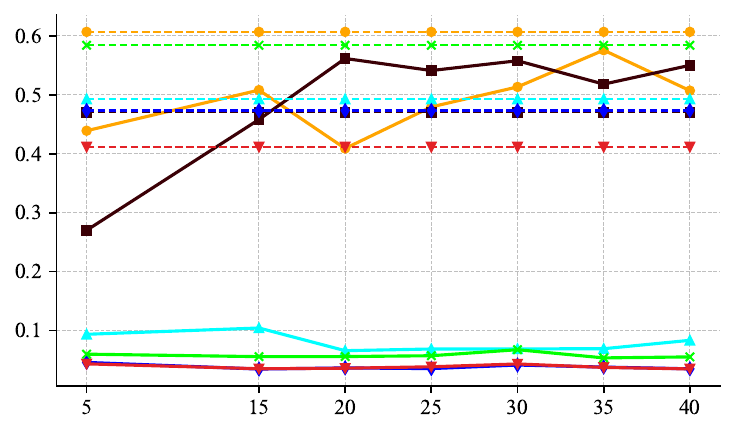} \\
    \small{RMM, URSS} & \small{RHL, SFD} & \small{RHL, URSS}\\
    \includegraphics[width=0.3\textwidth]{ 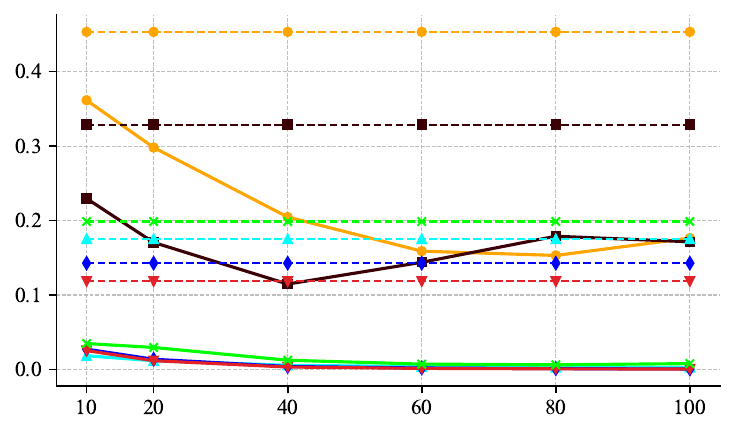} &
    \includegraphics[width=0.3\textwidth]{ 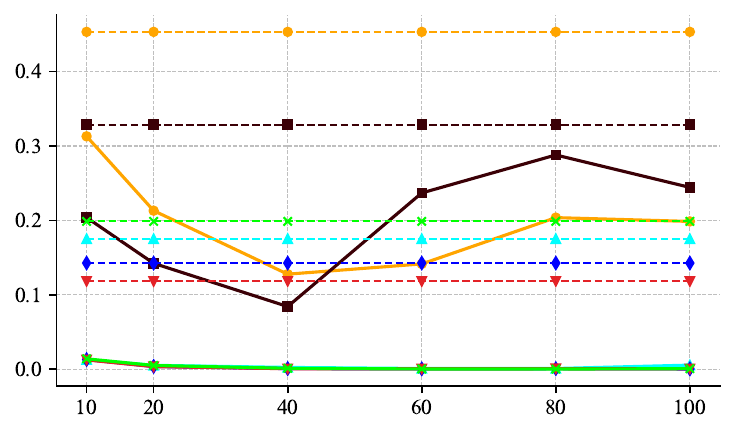}&
    \includegraphics[width=0.3\textwidth]{ 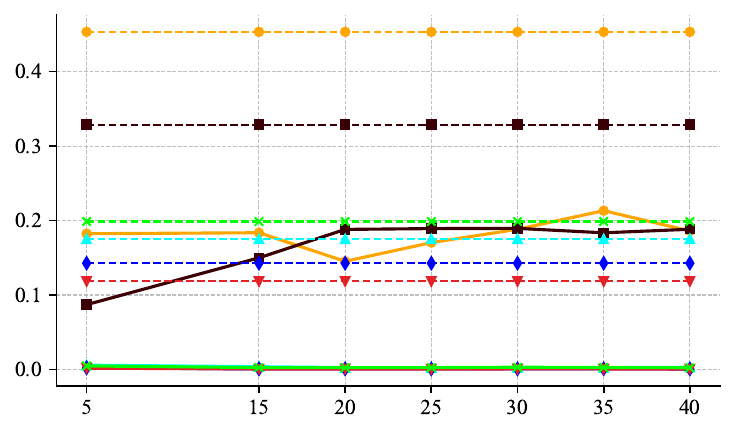}\\
    & \includegraphics[width=0.3\textwidth]{ 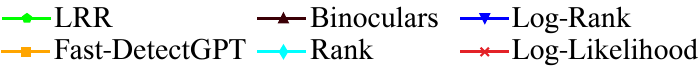} & \\
    \end{tabular}
    \caption{The performance of zero-shot detectors in RMM, AWS, and RHL strategies based on their hardness levels ($p\%$, $p\%$, and $R$, for RMM, AWS, and RHL, respectively).}
    \label{fig:hardness}
\end{figure*}

Figure \ref{fig:hardness} presents the trend of all evaluation metrics for each humanification strategy across their respective control parameters (i.e., $p$ ranging from $10\%$ to $100\%$ for RMM and AWS, and $R$ from $5$ to $40$ for RHL) for zero-shot detectors.
Dashed lines indicate the baseline, corresponding to paraphrased text without any humanification. 
Notably, both performance metric (W-AUROC) and the stability metric (SFD) consistently \textbf{fall below the baseline across all knob values}, confirming the effectiveness of the proposed strategies.
RHL achieves significant degradation in both W-AUROC and SFD starting from a low $R{\approx}15$, maintaining this effect across the full range.
AWS follows a similar trend at a moderate $p{\approx}60\%$, while RMM requires a higher $p{\approx}80\%$ to reach comparable impact.
This can be attributed to the fact that AWS and RHL deliberately replace high-entropy AI words with high-entropy human words, whereas RMM introduces more random substitutions, resulting in less targeted degradation.
A surprising observation is that some detectors exhibit increased stability as the hardness knob surpasses a certain threshold. 
This phenomenon is particularly pronounced in Binoculars and Fast-DetectGPT. 
A possible explanation is that as the text becomes more human-like, the discriminative signal diminishes (evidenced by lower W-AUROC), leading to more homogeneity across LLMs and writing styles. 
Consequently, detectors struggle to differentiate content and \textbf{converge to consistent decision thresholds}, resulting in reduced $\sigma_{\mathrm{FPR}}$.
While these systems exhibit improved consistency under intense adversarial conditions, their substantially degraded discriminative performance ultimately results in significantly penalized overall URSS scores.

An additional noteworthy observation derived from Figure \ref{fig:hardness} is the pronounced convergence of baseline trajectories in AUROC plots, which subsequently differentiates into distinctly separated lines across W-AUROC, SFD, and URSS plots.
This transformation from convergent to divergent patterns across different evaluation metrics provides further empirical validation for the conclusions presented in Section \ref{sec:why}, \textbf{why AUROC is not enough}, and illustrates how AUROC may offer misleadingly ``\textbf{easy wins}'' to certain detectors that may fail in \textbf{real-world applications}.

\section{Conclusion}
In this work, we presented \textbf{SHIELD}, a comprehensive benchmark designed to advance the fair evaluation of AI text detectors under realistic deployment scenarios.
Our results showed that conventional metrics like AUROC, despite their ubiquity, can significantly overestimate detector efficacy by neglecting critical operational constraints: 
the necessity of maintaining low FPR in practical AI text detection applications and the stability of detectors across varying threshold under deployment conditions where both writing style and generative model characteristics are typically unknown. 
\textbf{SHIELD} addresses these limitations through the introduction of USRR, a metric that integrates both performance measurement in low FPR regions and stability assessment.
Complementarily, \textbf{SHIELD} introduces a scalable humanification framework for generating humanified texts across graded difficulty levels, facilitating robust stress-testing of detection systems.
Our experimental findings reveal significant vulnerabilities in zero-shot detection methodologies, which exhibit performance degradation of approximately $80\%$ on average when subjected to even the most rudimentary word manipulation scenarios evaluated in this study, perturbations that closely approximate natural user editing behaviors without requiring specialized knowledge of AI- or human-authored text characteristics.

\section*{Limitations}
Several constraints merit acknowledgment within our experimental framework.
Primarily, our investigation was confined to monolingual English text analysis, precluding examination of multilingual detection scenarios that represent increasingly important deployment contexts given the global nature of AI text generation.
This linguistic limitation potentially restricts the generalizability of our findings to diverse language environments.
Additionally, the inherently dynamic nature of LLM development presents a significant temporal constraint; as generative architectures evolve through version updates and architectural innovations, their statistical signatures undergo corresponding transformations, necessitating periodic collection of representative text samples to maintain benchmark currency and relevance. 
Furthermore, budgetary constraints confined our experimental protocol to open-source models, resulting in the exclusion of closed-source systems including ChatGPT and Claude. 
The inclusion of these commercial platforms would enhance the comprehensiveness of our evaluation framework, particularly considering their extensive deployment and potentially distinctive generative characteristics that may present unique challenges to detection methodologies.

\section*{Ethical considerations}
This investigation aims to evaluate the robustness of contemporary AI-text detection methods against adversarial manipulations. The proliferation of LLM-generated content and its potential for malicious applications necessitates robust detection mechanisms to serve as effective countermeasures against synthetic text deception. Vulnerabilities in these detection frameworks could precipitate significant complications in computational forensics and information verification processes. Consequently, this research endeavors to provide detection system engineers with rigorous adversarial testing frameworks for comprehensive validation of their algorithmic approaches against sophisticated evasion techniques. We explicitly stipulate that the methodologies and results documented in this study are intended exclusively for detection system improvement and validation, and not for circumvention of existing detection systems.

\bibliography{custom}

\begin{thebibliography}{55}
\providecommand{\natexlab}[1]{#1}

\bibitem[{Abassy et~al.(2024)Abassy, Elozeiri, Aziz, Ta, Tomar, Adhikari, Ahmed, Wang, Mohammed~Afzal, Xie, Mansurov, Artemova, Mikhailov, Xing, Geng, Iqbal, Mujahid, Mahmoud, Tsvigun, Aji, Shelmanov, Habash, Gurevych, and Nakov}]{abassy-etal-2024-llm}
Mervat Abassy, Kareem Elozeiri, Alexander Aziz, Minh~Ngoc Ta, Raj~Vardhan Tomar, Bimarsha Adhikari, Saad El~Dine Ahmed, Yuxia Wang, Osama Mohammed~Afzal, Zhuohan Xie, Jonibek Mansurov, Ekaterina Artemova, Vladislav Mikhailov, Rui Xing, Jiahui Geng, Hasan Iqbal, Zain~Muhammad Mujahid, Tarek Mahmoud, Akim Tsvigun, and 5 others. 2024.
\newblock \href {https://doi.org/10.18653/v1/2024.emnlp-demo.35} {{LLM}-{D}etect{AI}ve: a tool for fine-grained machine-generated text detection}.
\newblock In \emph{Proceedings of the 2024 Conference on Empirical Methods in Natural Language Processing: System Demonstrations}, pages 336--343, Miami, Florida, USA. Association for Computational Linguistics.

\bibitem[{Ayoobi et~al.(2024)Ayoobi, Knab, Cheng, Pantoja, Alikhani, Flamant, Kim, and Mukherjee}]{ayoobi2024esperanto}
Navid Ayoobi, Lily Knab, Wen Cheng, David Pantoja, Hamidreza Alikhani, Sylvain Flamant, Jin Kim, and Arjun Mukherjee. 2024.
\newblock Esperanto: Evaluating synthesized phrases to enhance robustness in ai detection for text origination.
\newblock \emph{arXiv preprint arXiv:2409.14285}.

\bibitem[{Bai et~al.(2023)Bai, Bai, Chu, Cui, Dang, Deng, Fan, Ge, Han, Huang et~al.}]{bai2023qwen}
Jinze Bai, Shuai Bai, Yunfei Chu, Zeyu Cui, Kai Dang, Xiaodong Deng, Yang Fan, Wenbin Ge, Yu~Han, Fei Huang, and 1 others. 2023.
\newblock Qwen technical report.
\newblock \emph{arXiv preprint arXiv:2309.16609}.

\bibitem[{Bao et~al.(2023)Bao, Zhao, Teng, Yang, and Zhang}]{bao2023fast}
Guangsheng Bao, Yanbin Zhao, Zhiyang Teng, Linyi Yang, and Yue Zhang. 2023.
\newblock Fast-detectgpt: Efficient zero-shot detection of machine-generated text via conditional probability curvature.
\newblock \emph{arXiv preprint arXiv:2310.05130}.

\bibitem[{Beltagy et~al.(2020)Beltagy, Peters, and Cohan}]{Beltagy2020Longformer}
Iz~Beltagy, Matthew~E. Peters, and Arman Cohan. 2020.
\newblock Longformer: The long-document transformer.
\newblock \emph{arXiv:2004.05150}.

\bibitem[{Chiang et~al.(2023)Chiang, Chen, Song, Shuai, and Chang}]{10.1145/3580305.3599502}
Hung-Yun Chiang, Yi-Syuan Chen, Yun-Zhu Song, Hong-Han Shuai, and Jason~S. Chang. 2023.
\newblock \href {https://doi.org/10.1145/3580305.3599502} {Shilling black-box review-based recommender systems through fake review generation}.
\newblock In \emph{Proceedings of the 29th ACM SIGKDD Conference on Knowledge Discovery and Data Mining}, KDD '23, page 286–297, New York, NY, USA. Association for Computing Machinery.

\bibitem[{Clark et~al.(2021)Clark, August, Serrano, Haduong, Gururangan, and Smith}]{clark-etal-2021-thats}
Elizabeth Clark, Tal August, Sofia Serrano, Nikita Haduong, Suchin Gururangan, and Noah~A. Smith. 2021.
\newblock \href {https://doi.org/10.18653/v1/2021.acl-long.565} {All that`s {\textquoteleft}human' is not gold: Evaluating human evaluation of generated text}.
\newblock In \emph{Proceedings of the 59th Annual Meeting of the Association for Computational Linguistics and the 11th International Joint Conference on Natural Language Processing (Volume 1: Long Papers)}, pages 7282--7296, Online. Association for Computational Linguistics.

\bibitem[{Clement et~al.(2019)Clement, Bierbaum, O'Keeffe, and Alemi}]{clement2019use}
Colin~B Clement, Matthew Bierbaum, Kevin~P O'Keeffe, and Alexander~A Alemi. 2019.
\newblock On the use of arxiv as a dataset.
\newblock \emph{arXiv preprint arXiv:1905.00075}.

\bibitem[{Cornelius et~al.(2024)Cornelius, Lithgow-Serrano, Mitrovic, Dolamic, and Rinaldi}]{cornelius-etal-2024-bust}
Joseph Cornelius, Oscar Lithgow-Serrano, Sandra Mitrovic, Ljiljana Dolamic, and Fabio Rinaldi. 2024.
\newblock \href {https://doi.org/10.18653/v1/2024.naacl-long.444} {{BUST}: Benchmark for the evaluation of detectors of {LLM}-generated text}.
\newblock In \emph{Proceedings of the 2024 Conference of the North American Chapter of the Association for Computational Linguistics: Human Language Technologies (Volume 1: Long Papers)}, pages 8029--8057, Mexico City, Mexico. Association for Computational Linguistics.

\bibitem[{Cotton et~al.(2024)Cotton, Cotton, and Shipway}]{Cotton03032024}
Debby R.~E. Cotton, Peter~A. Cotton, and J.~Reuben Shipway. 2024.
\newblock \href {https://doi.org/10.1080/14703297.2023.2190148} {Chatting and cheating: Ensuring academic integrity in the era of chatgpt}.
\newblock \emph{Innovations in Education and Teaching International}, 61(2):228--239.

\bibitem[{Dou et~al.(2022)Dou, Forbes, Koncel-Kedziorski, Smith, and Choi}]{dou-etal-2022-gpt}
Yao Dou, Maxwell Forbes, Rik Koncel-Kedziorski, Noah~A. Smith, and Yejin Choi. 2022.
\newblock \href {https://doi.org/10.18653/v1/2022.acl-long.501} {Is {GPT}-3 text indistinguishable from human text? scarecrow: A framework for scrutinizing machine text}.
\newblock In \emph{Proceedings of the 60th Annual Meeting of the Association for Computational Linguistics (Volume 1: Long Papers)}, pages 7250--7274, Dublin, Ireland. Association for Computational Linguistics.

\bibitem[{Dugan et~al.(2024)Dugan, Hwang, Trhl{\'i}k, Zhu, Ludan, Xu, Ippolito, and Callison-Burch}]{dugan2024raid}
Liam Dugan, Alyssa Hwang, Filip Trhl{\'i}k, Andrew Zhu, Josh~Magnus Ludan, Hainiu Xu, Daphne Ippolito, and Chris Callison-Burch. 2024.
\newblock \href {https://doi.org/10.18653/v1/2024.acl-long.674} {{RAID}: A shared benchmark for robust evaluation of machine-generated text detectors}.
\newblock In \emph{Proceedings of the 62nd Annual Meeting of the Association for Computational Linguistics (Volume 1: Long Papers)}, pages 12463--12492, Bangkok, Thailand. Association for Computational Linguistics.

\bibitem[{Fan et~al.(2019)Fan, Jernite, Perez, Grangier, Weston, and Auli}]{fan-etal-2019-eli5}
Angela Fan, Yacine Jernite, Ethan Perez, David Grangier, Jason Weston, and Michael Auli. 2019.
\newblock \href {https://doi.org/10.18653/v1/P19-1346} {{ELI}5: Long form question answering}.
\newblock In \emph{Proceedings of the 57th Annual Meeting of the Association for Computational Linguistics}, pages 3558--3567, Florence, Italy. Association for Computational Linguistics.

\bibitem[{Gehrmann et~al.(2019)Gehrmann, Strobelt, and Rush}]{gehrmann-etal-2019-gltr}
Sebastian Gehrmann, Hendrik Strobelt, and Alexander Rush. 2019.
\newblock \href {https://doi.org/10.18653/v1/P19-3019} {{GLTR}: Statistical detection and visualization of generated text}.
\newblock In \emph{Proceedings of the 57th Annual Meeting of the Association for Computational Linguistics: System Demonstrations}, pages 111--116, Florence, Italy. Association for Computational Linguistics.

\bibitem[{Grattafiori et~al.(2024)Grattafiori, Dubey, Jauhri, Pandey, Kadian, Al-Dahle, Letman, Mathur, Schelten, Vaughan et~al.}]{grattafiori2024llama}
Aaron Grattafiori, Abhimanyu Dubey, Abhinav Jauhri, Abhinav Pandey, Abhishek Kadian, Ahmad Al-Dahle, Aiesha Letman, Akhil Mathur, Alan Schelten, Alex Vaughan, and 1 others. 2024.
\newblock The llama 3 herd of models.
\newblock \emph{arXiv preprint arXiv:2407.21783}.

\bibitem[{Guo et~al.(2023)Guo, Zhang, Wang, Jiang, Nie, Ding, Yue, and Wu}]{guo2023close}
Biyang Guo, Xin Zhang, Ziyuan Wang, Minqi Jiang, Jinran Nie, Yuxuan Ding, Jianwei Yue, and Yupeng Wu. 2023.
\newblock How close is chatgpt to human experts? comparison corpus, evaluation, and detection.
\newblock \emph{arXiv preprint arXiv:2301.07597}.

\bibitem[{Guo et~al.(2024{\natexlab{a}})Guo, Cheng, Jin, Zhang, Zhang, Tao, Shen, and Zhang}]{guo2024biscope}
Hanxi Guo, Siyuan Cheng, Xiaolong Jin, Zhuo Zhang, Kaiyuan Zhang, Guanhong Tao, Guangyu Shen, and Xiangyu Zhang. 2024{\natexlab{a}}.
\newblock \href {https://proceedings.neurips.cc/paper_files/paper/2024/file/bc808cf2d2444b0abcceca366b771389-Paper-Conference.pdf} {Biscope: Ai-generated text detection by checking memorization of preceding tokens}.
\newblock In \emph{Advances in Neural Information Processing Systems}, volume~37, pages 104065--104090. Curran Associates, Inc.

\bibitem[{Guo et~al.(2024{\natexlab{b}})Guo, Zhang, He, Zhang, Feng, Huang, and Ma}]{guo2024detective}
Xun Guo, Shan Zhang, Yongxin He, Ting Zhang, Wanquan Feng, Haibin Huang, and Chongyang Ma. 2024{\natexlab{b}}.
\newblock \href {https://proceedings.neurips.cc/paper_files/paper/2024/file/a117a3cd54b7affad04618c77c2fb18b-Paper-Conference.pdf} {Detective: Detecting ai-generated text via multi-level contrastive learning}.
\newblock In \emph{Advances in Neural Information Processing Systems}, volume~37, pages 88320--88347. Curran Associates, Inc.

\bibitem[{Hans et~al.(2024)Hans, Schwarzschild, Cherepanova, Kazemi, Saha, Goldblum, Geiping, and Goldstein}]{hans2024spotting}
Abhimanyu Hans, Avi Schwarzschild, Valeriia Cherepanova, Hamid Kazemi, Aniruddha Saha, Micah Goldblum, Jonas Geiping, and Tom Goldstein. 2024.
\newblock Spotting {LLM}s with binoculars: zero-shot detection of machine-generated text.
\newblock In \emph{Proceedings of the 41st International Conference on Machine Learning}, ICML'24. JMLR.org.

\bibitem[{He et~al.(2024)He, Shen, Chen, Backes, and Zhang}]{he2024mgtbench}
Xinlei He, Xinyue Shen, Zeyuan Chen, Michael Backes, and Yang Zhang. 2024.
\newblock \href {https://doi.org/10.1145/3658644.3670344} {Mgtbench: Benchmarking machine-generated text detection}.
\newblock In \emph{Proceedings of the 2024 on ACM SIGSAC Conference on Computer and Communications Security}, CCS '24, page 2251–2265, New York, NY, USA. Association for Computing Machinery.

\bibitem[{Horne and Gruppi(2024)}]{Horne_Gruppi_2024}
Benjamin~D. Horne and Maurício Gruppi. 2024.
\newblock \href {https://doi.org/10.1609/icwsm.v18i1.31439} {Nela-ps: A dataset of pink slime news articles for the study of local news ecosystems}.
\newblock \emph{Proceedings of the International AAAI Conference on Web and Social Media}, 18(1):1958--1966.

\bibitem[{Hou et~al.(2024)Hou, Shen, and Lu}]{hou-etal-2024-progressive}
Guiyang Hou, Yongliang Shen, and Weiming Lu. 2024.
\newblock \href {https://doi.org/10.18653/v1/2024.findings-acl.855} {Progressive tuning: Towards generic sentiment abilities for large language models}.
\newblock In \emph{Findings of the Association for Computational Linguistics: ACL 2024}, pages 14392--14402, Bangkok, Thailand. Association for Computational Linguistics.

\bibitem[{Hu et~al.(2023)Hu, Chen, and Ho}]{NEURIPS2023_30e15e59}
Xiaomeng Hu, Pin-Yu Chen, and Tsung-Yi Ho. 2023.
\newblock \href {https://proceedings.neurips.cc/paper_files/paper/2023/file/30e15e5941ae0cdab7ef58cc8d59a4ca-Paper-Conference.pdf} {Radar: Robust ai-text detection via adversarial learning}.
\newblock In \emph{Advances in Neural Information Processing Systems}, volume~36, pages 15077--15095. Curran Associates, Inc.

\bibitem[{Huang et~al.(2024)Huang, Zhang, Li, You, Wang, and Yang}]{huang-etal-2024-ai}
Guanhua Huang, Yuchen Zhang, Zhe Li, Yongjian You, Mingze Wang, and Zhouwang Yang. 2024.
\newblock \href {https://doi.org/10.18653/v1/2024.acl-long.327} {Are {AI}-generated text detectors robust to adversarial perturbations?}
\newblock In \emph{Proceedings of the 62nd Annual Meeting of the Association for Computational Linguistics (Volume 1: Long Papers)}, pages 6005--6024, Bangkok, Thailand. Association for Computational Linguistics.

\bibitem[{Jiang et~al.(2023)Jiang, Sablayrolles, Mensch, Bamford, Chaplot, de~las Casas, Bressand, Lengyel, Lample, Saulnier, Lavaud, Lachaux, Stock, Scao, Lavril, Wang, Lacroix, and Sayed}]{jiang2023mistral7b}
Albert~Q. Jiang, Alexandre Sablayrolles, Arthur Mensch, Chris Bamford, Devendra~Singh Chaplot, Diego de~las Casas, Florian Bressand, Gianna Lengyel, Guillaume Lample, Lucile Saulnier, Lélio~Renard Lavaud, Marie-Anne Lachaux, Pierre Stock, Teven~Le Scao, Thibaut Lavril, Thomas Wang, Timothée Lacroix, and William~El Sayed. 2023.
\newblock \href {https://arxiv.org/abs/2310.06825} {Mistral 7b}.
\newblock \emph{Preprint}, arXiv:2310.06825.

\bibitem[{Kamalloo et~al.(2023)Kamalloo, Dziri, Clarke, and Rafiei}]{kamalloo-etal-2023-evaluating}
Ehsan Kamalloo, Nouha Dziri, Charles Clarke, and Davood Rafiei. 2023.
\newblock \href {https://doi.org/10.18653/v1/2023.acl-long.307} {Evaluating open-domain question answering in the era of large language models}.
\newblock In \emph{Proceedings of the 61st Annual Meeting of the Association for Computational Linguistics (Volume 1: Long Papers)}, pages 5591--5606, Toronto, Canada. Association for Computational Linguistics.

\bibitem[{Kirchenbauer et~al.(2023)Kirchenbauer, Geiping, Wen, Katz, Miers, and Goldstein}]{pmlr-v202-kirchenbauer23a}
John Kirchenbauer, Jonas Geiping, Yuxin Wen, Jonathan Katz, Ian Miers, and Tom Goldstein. 2023.
\newblock \href {https://proceedings.mlr.press/v202/kirchenbauer23a.html} {A watermark for large language models}.
\newblock In \emph{Proceedings of the 40th International Conference on Machine Learning}, volume 202 of \emph{Proceedings of Machine Learning Research}, pages 17061--17084. PMLR.

\bibitem[{Kuznetsov et~al.(2024)Kuznetsov, Tulchinskii, Kushnareva, Magai, Barannikov, Nikolenko, and Piontkovskaya}]{kuznetsov-etal-2024-robust}
Kristian Kuznetsov, Eduard Tulchinskii, Laida Kushnareva, German Magai, Serguei Barannikov, Sergey Nikolenko, and Irina Piontkovskaya. 2024.
\newblock \href {https://doi.org/10.18653/v1/2024.findings-emnlp.992} {Robust {AI}-generated text detection by restricted embeddings}.
\newblock In \emph{Findings of the Association for Computational Linguistics: EMNLP 2024}, pages 17036--17055, Miami, Florida, USA. Association for Computational Linguistics.

\bibitem[{Lai et~al.(2024)Lai, Zhang, and Chen}]{10651296}
Zhixin Lai, Xuesheng Zhang, and Suiyao Chen. 2024.
\newblock \href {https://doi.org/10.1109/IJCNN60899.2024.10651296} {Adaptive ensembles of fine-tuned transformers for {LLM}-generated text detection}.
\newblock In \emph{2024 International Joint Conference on Neural Networks (IJCNN)}, pages 1--7.

\bibitem[{Lavergne et~al.(2008)Lavergne, Urvoy, and Yvon}]{10.5555/3053718.3053722}
Thomas Lavergne, Tanguy Urvoy, and Fran\c{c}ois Yvon. 2008.
\newblock Detecting fake content with relative entropy scoring.
\newblock In \emph{Proceedings of the 2008 International Conference on Uncovering Plagiarism, Authorship and Social Software Misuse - Volume 377}, PAN'08, page 27–31, Aachen, DEU. CEUR-WS.org.

\bibitem[{Li et~al.(2024)Li, Li, Cui, Bi, Wang, Wang, Yang, Shi, and Zhang}]{li2024mage}
Yafu Li, Qintong Li, Leyang Cui, Wei Bi, Zhilin Wang, Longyue Wang, Linyi Yang, Shuming Shi, and Yue Zhang. 2024.
\newblock \href {https://doi.org/10.18653/v1/2024.acl-long.3} {{MAGE}: Machine-generated text detection in the wild}.
\newblock In \emph{Proceedings of the 62nd Annual Meeting of the Association for Computational Linguistics (Volume 1: Long Papers)}, pages 36--53, Bangkok, Thailand. Association for Computational Linguistics.

\bibitem[{Liu and Bu(2024)}]{liu2024adaptive}
Yepeng Liu and Yuheng Bu. 2024.
\newblock Adaptive text watermark for large language models.
\newblock In \emph{Proceedings of the 41st International Conference on Machine Learning}, ICML'24. JMLR.org.

\bibitem[{Liu et~al.(2021)Liu, Lin, Shi, and Zhao}]{10.1007/978-3-030-84186-7_31}
Zhuang Liu, Wayne Lin, Ya~Shi, and Jun Zhao. 2021.
\newblock A robustly optimized bert pre-training approach with post-training.
\newblock In \emph{Chinese Computational Linguistics}, pages 471--484, Cham. Springer International Publishing.

\bibitem[{Ma and Wang(2024)}]{ma-wang-2024-zero}
Shixuan Ma and Quan Wang. 2024.
\newblock \href {https://doi.org/10.18653/v1/2024.emnlp-main.971} {Zero-shot detection of {LLM}-generated text using token cohesiveness}.
\newblock In \emph{Proceedings of the 2024 Conference on Empirical Methods in Natural Language Processing}, pages 17538--17553, Miami, Florida, USA. Association for Computational Linguistics.

\bibitem[{Mesnard et~al.(2024)Mesnard, Hardin, Dadashi, Bhupatiraju, Pathak, Sifre, Rivi{\`e}re, Kale, Love et~al.}]{team2024gemma}
Thomas Mesnard, Cassidy Hardin, Robert Dadashi, Surya Bhupatiraju, Shreya Pathak, Laurent Sifre, Morgane Rivi{\`e}re, Mihir~Sanjay Kale, Juliette Love, and 1 others. 2024.
\newblock Gemma: Open models based on gemini research and technology.
\newblock \emph{arXiv preprint arXiv:2403.08295}.

\bibitem[{Mitchell et~al.(2023)Mitchell, Lee, Khazatsky, Manning, and Finn}]{10.5555/3618408.3619446}
Eric Mitchell, Yoonho Lee, Alexander Khazatsky, Christopher~D. Manning, and Chelsea Finn. 2023.
\newblock Detectgpt: zero-shot machine-generated text detection using probability curvature.
\newblock In \emph{Proceedings of the 40th International Conference on Machine Learning}, ICML'23. JMLR.org.

\bibitem[{Ni et~al.(2019)Ni, Li, and McAuley}]{ni-etal-2019-justifying}
Jianmo Ni, Jiacheng Li, and Julian McAuley. 2019.
\newblock \href {https://doi.org/10.18653/v1/D19-1018} {Justifying recommendations using distantly-labeled reviews and fine-grained aspects}.
\newblock In \emph{Proceedings of the 2019 Conference on Empirical Methods in Natural Language Processing and the 9th International Joint Conference on Natural Language Processing (EMNLP-IJCNLP)}, pages 188--197, Hong Kong, China. Association for Computational Linguistics.

\bibitem[{Panaitescu-Liess et~al.(2025)Panaitescu-Liess, Che, An, Xu, Pathmanathan, Chakraborty, Zhu, Goldstein, and Huang}]{Panaitescu-Liess_Che_An_Xu_Pathmanathan_Chakraborty_Zhu_Goldstein_Huang_2025}
Michael-Andrei Panaitescu-Liess, Zora Che, Bang An, Yuancheng Xu, Pankayaraj Pathmanathan, Souradip Chakraborty, Sicheng Zhu, Tom Goldstein, and Furong Huang. 2025.
\newblock \href {https://doi.org/10.1609/aaai.v39i23.34684} {Can watermarking large language models prevent copyrighted text generation and hide training data?}
\newblock \emph{Proceedings of the AAAI Conference on Artificial Intelligence}, 39(23):25002--25009.

\bibitem[{Pang et~al.(2024)Pang, Hu, Zheng, and Smith}]{pang2024no}
Qi~Pang, Shengyuan Hu, Wenting Zheng, and Virginia Smith. 2024.
\newblock No free lunch in {LLM} watermarking: Trade-offs in watermarking design choices.
\newblock \emph{arXiv preprint arXiv:2402.16187}.

\bibitem[{Pudasaini et~al.(2025)Pudasaini, Miralles, Lillis, and Salvador}]{pudasaini-etal-2025-benchmarking}
Shushanta Pudasaini, Luis Miralles, David Lillis, and Marisa~Llorens Salvador. 2025.
\newblock \href {https://aclanthology.org/2025.genaidetect-1.4/} {Benchmarking {AI} text detection: Assessing detectors against new datasets, evasion tactics, and enhanced {LLM}s}.
\newblock In \emph{Proceedings of the 1stWorkshop on GenAI Content Detection (GenAIDetect)}, pages 68--77, Abu Dhabi, UAE. International Conference on Computational Linguistics.

\bibitem[{Rawte et~al.(2023)Rawte, Chakraborty, Pathak, Sarkar, Tonmoy, Chadha, Sheth, and Das}]{rawte-etal-2023-troubling}
Vipula Rawte, Swagata Chakraborty, Agnibh Pathak, Anubhav Sarkar, S.M Towhidul~Islam Tonmoy, Aman Chadha, Amit Sheth, and Amitava Das. 2023.
\newblock \href {https://doi.org/10.18653/v1/2023.emnlp-main.155} {The troubling emergence of hallucination in large language models - an extensive definition, quantification, and prescriptive remediations}.
\newblock In \emph{Proceedings of the 2023 Conference on Empirical Methods in Natural Language Processing}, pages 2541--2573, Singapore. Association for Computational Linguistics.

\bibitem[{Su et~al.(2023)Su, Zhuo, Wang, and Nakov}]{su2023detectllm}
Jinyan Su, Terry Zhuo, Di~Wang, and Preslav Nakov. 2023.
\newblock \href {https://doi.org/10.18653/v1/2023.findings-emnlp.827} {{D}etect{LLM}: Leveraging log rank information for zero-shot detection of machine-generated text}.
\newblock In \emph{Findings of the Association for Computational Linguistics: EMNLP 2023}, pages 12395--12412, Singapore. Association for Computational Linguistics.

\bibitem[{Uchendu et~al.(2021)Uchendu, Ma, Le, Zhang, and Lee}]{uchendu-etal-2021-turingbench-benchmark}
Adaku Uchendu, Zeyu Ma, Thai Le, Rui Zhang, and Dongwon Lee. 2021.
\newblock \href {https://doi.org/10.18653/v1/2021.findings-emnlp.172} {{TURINGBENCH}: A benchmark environment for {T}uring test in the age of neural text generation}.
\newblock In \emph{Findings of the Association for Computational Linguistics: EMNLP 2021}, pages 2001--2016, Punta Cana, Dominican Republic. Association for Computational Linguistics.

\bibitem[{Vykopal et~al.(2024)Vykopal, Pikuliak, Srba, Moro, Macko, and Bielikova}]{vykopal-etal-2024-disinformation}
Ivan Vykopal, Mat{\'u}{\v{s}} Pikuliak, Ivan Srba, Robert Moro, Dominik Macko, and Maria Bielikova. 2024.
\newblock \href {https://doi.org/10.18653/v1/2024.acl-long.793} {Disinformation capabilities of large language models}.
\newblock In \emph{Proceedings of the 62nd Annual Meeting of the Association for Computational Linguistics (Volume 1: Long Papers)}, pages 14830--14847, Bangkok, Thailand. Association for Computational Linguistics.

\bibitem[{Wahle et~al.(2022)Wahle, Ruas, Kirstein, and Gipp}]{wahle-etal-2022-large}
Jan~Philip Wahle, Terry Ruas, Frederic Kirstein, and Bela Gipp. 2022.
\newblock \href {https://doi.org/10.18653/v1/2022.emnlp-main.62} {How large language models are transforming machine-paraphrase plagiarism}.
\newblock In \emph{Proceedings of the 2022 Conference on Empirical Methods in Natural Language Processing}, pages 952--963, Abu Dhabi, United Arab Emirates. Association for Computational Linguistics.

\bibitem[{Wang et~al.(2023)Wang, Zhang, and Wang}]{wang-etal-2023-element}
Yiming Wang, Zhuosheng Zhang, and Rui Wang. 2023.
\newblock \href {https://doi.org/10.18653/v1/2023.acl-long.482} {Element-aware summarization with large language models: Expert-aligned evaluation and chain-of-thought method}.
\newblock In \emph{Proceedings of the 61st Annual Meeting of the Association for Computational Linguistics (Volume 1: Long Papers)}, pages 8640--8665, Toronto, Canada. Association for Computational Linguistics.

\bibitem[{Wang et~al.(2024)Wang, Mansurov, Ivanov, Su, Shelmanov, Tsvigun, Mohammed~Afzal, Mahmoud, Puccetti, Arnold, Aji, Habash, Gurevych, and Nakov}]{wang2024m4gt}
Yuxia Wang, Jonibek Mansurov, Petar Ivanov, Jinyan Su, Artem Shelmanov, Akim Tsvigun, Osama Mohammed~Afzal, Tarek Mahmoud, Giovanni Puccetti, Thomas Arnold, Alham Aji, Nizar Habash, Iryna Gurevych, and Preslav Nakov. 2024.
\newblock \href {https://doi.org/10.18653/v1/2024.acl-long.218} {{M}4{GT}-bench: Evaluation benchmark for black-box machine-generated text detection}.
\newblock In \emph{Proceedings of the 62nd Annual Meeting of the Association for Computational Linguistics (Volume 1: Long Papers)}, pages 3964--3992, Bangkok, Thailand. Association for Computational Linguistics.

\bibitem[{Wu et~al.(2025)Wu, Yang, Zhan, Yuan, Chao, and Wong}]{10.1162/coli_a_00549}
Junchao Wu, Shu Yang, Runzhe Zhan, Yulin Yuan, Lidia~Sam Chao, and Derek~Fai Wong. 2025.
\newblock \href {https://doi.org/10.1162/coli_a_00549} {A survey on {LLM}-generated text detection: Necessity, methods, and future directions}.
\newblock \emph{Computational Linguistics}, 51(1):275--338.

\bibitem[{Wu et~al.(2024)Wu, Zhan, Wong, Yang, Yang, Yuan, and Chao}]{wu2024detectrl}
Junchao Wu, Runzhe Zhan, Derek~F. Wong, Shu Yang, Xinyi Yang, Yulin Yuan, and Lidia~S. Chao. 2024.
\newblock \href {https://proceedings.neurips.cc/paper_files/paper/2024/file/b61bdf7e9f64c04ec75a26e781e2ad51-Paper-Datasets_and_Benchmarks_Track.pdf} {Detectrl: Benchmarking {LLM}-generated text detection in real-world scenarios}.
\newblock In \emph{Advances in Neural Information Processing Systems}, volume~37, pages 100369--100401. Curran Associates, Inc.

\bibitem[{Yang et~al.(2023)Yang, Cheng, Wu, Petzold, Wang, and Chen}]{yang2023dna}
Xianjun Yang, Wei Cheng, Yue Wu, Linda Petzold, William~Yang Wang, and Haifeng Chen. 2023.
\newblock Dna-gpt: Divergent n-gram analysis for training-free detection of gpt-generated text.
\newblock \emph{arXiv preprint arXiv:2305.17359}.

\bibitem[{Youden(1950)}]{youden1950index}
WJ~Youden. 1950.
\newblock Index for rating diagnostic tests.
\newblock \emph{Cancer}, 3(1):32--35.

\bibitem[{Yu et~al.(2025)Yu, Luo, Madusu, Lal, and Howard}]{yu2025your}
Sungduk Yu, Man Luo, Avinash Madusu, Vasudev Lal, and Phillip Howard. 2025.
\newblock Is your paper being reviewed by an {LLM}? a new benchmark dataset and approach for detecting ai text in peer review.
\newblock \emph{arXiv preprint arXiv:2502.19614}.

\bibitem[{Yu et~al.(2024{\natexlab{a}})Yu, Chen, Yang, Zhang, and Yu}]{yu-etal-2024-text}
Xiao Yu, Kejiang Chen, Qi~Yang, Weiming Zhang, and Nenghai Yu. 2024{\natexlab{a}}.
\newblock \href {https://doi.org/10.18653/v1/2024.emnlp-main.885} {Text fluoroscopy: Detecting {LLM}-generated text through intrinsic features}.
\newblock In \emph{Proceedings of the 2024 Conference on Empirical Methods in Natural Language Processing}, pages 15838--15846, Miami, Florida, USA. Association for Computational Linguistics.

\bibitem[{Yu et~al.(2024{\natexlab{b}})Yu, Qi, Chen, Chen, Yang, Zhu, Shang, Zhang, and Yu}]{NEURIPS2024_1d35af80}
Xiao Yu, Yuang Qi, Kejiang Chen, Guoqiang Chen, Xi~Yang, Pengyuan Zhu, Xiuwei Shang, Weiming Zhang, and Nenghai Yu. 2024{\natexlab{b}}.
\newblock \href {https://proceedings.neurips.cc/paper_files/paper/2024/file/1d35af80e775e342f4cd3792e4405837-Paper-Conference.pdf} {Dpic: Decoupling prompt and intrinsic characteristics for {LLM} generated text detection}.
\newblock In \emph{Advances in Neural Information Processing Systems}, volume~37, pages 16194--16212. Curran Associates, Inc.

\bibitem[{Zhou et~al.(2024)Zhou, He, and Sun}]{zhou-etal-2024-humanizing}
Ying Zhou, Ben He, and Le~Sun. 2024.
\newblock \href {https://aclanthology.org/2024.lrec-main.739/} {Humanizing machine-generated content: Evading {AI}-text detection through adversarial attack}.
\newblock In \emph{Proceedings of the 2024 Joint International Conference on Computational Linguistics, Language Resources and Evaluation (LREC-COLING 2024)}, pages 8427--8437, Torino, Italia. ELRA and ICCL.

\end{thebibliography}
\clearpage

\appendix
\section{Dataset}\label{app:dataset}
\subsection{Data domains}\label{app:data}
\noindent
\textbf{Medium:}
We utilized the Medium Articles dataset curated by Fabio Chiusano, hosted on Hugging Face \footnote{\url{https://huggingface.co/datasets/fabiochiu/medium-articles}}.
This dataset consists of more than 190k English-language articles from Medium.com, each containing textual and metadata fields, including title, body, URL, authorship, timestamp, and tags.
We randomly selected a subset of 12.5k articles with word counts between 400 and 2000 that were published before 2021.

\noindent
\textbf{News:}
We curated a collection of 200k news articles from reputable news agency websites, including ABC News, Al Jazeera, American Press, Associated Press News, CBS News, CNN, NBC News, Reuters, and The Guardian. From this corpus, we randomly sampled 12.5k articles with lengths ranging from 250 to 2000 words. The selected articles span diverse topical domains such as politics, science, social issues, religion, technology, sports, and culture. All articles were published prior to the advent of modern LLMs.

\noindent
\textbf{Reviews:}
We utilized the Amazon Reviews dataset collected by \cite{ni-etal-2019-justifying}, which comprises over 233 million customer reviews spanning from 1996 to October 2018. This large-scale dataset includes rich metadata, such as review text, star ratings, helpfulness scores, and product attributes (e.g., category, brand, price, and image features). For our benchmark, we randomly sampled 12.5k reviews from various product categories, retaining only those with 30 or more words.

\noindent
\textbf{Reddit:}
To build the Reddit component of our dataset, we extracted question–answer pairs from the ELI5 subreddit, following a collection strategy similar to \cite{fan-etal-2019-eli5}. We restricted the dataset to answers with lengths between 400 and 2000 words, all posted before 2021. A final sample of 12.5k answers was randomly selected for inclusion in our benchmark.

\noindent
\textbf{arXiv:}
We utilized the arXiv dataset introduced by \cite{clement2019use}, which contains over 1.5 million preprint articles spanning disciplines such as physics, mathematics, and computer science. 
Each article includes metadata such as title, abstract, authors, categories, and citation data. 
To construct our dataset, we randomly sampled 12.5k abstracts with word counts ranging from 150 to 500, limited to publications before 2021.

\noindent
\textbf{Pink slime:}
We employed the NELA-PS dataset introduced by \cite{Horne_Gruppi_2024}, which encompasses 7.9 million articles from 1093 local news sources, commonly referred to as ``pink slime'' journalism, spanning from March 2021 to January 2024. These outlets generate content that mimics legitimate local journalism in structure and presentation while frequently advancing partisan narratives. 
From this corpus, we sampled 12.5k articles published in 2021, with document lengths constrained to 250-2000 words.

\noindent
\textbf{Wikipedia}
We utilized the Plain Text Wikipedia 2020-11 dataset accessible through Kaggle \footnote{\url{https://www.kaggle.com/datasets/ltcmdrdata/plain-text-wikipedia-202011}}. This corpus comprises a comprehensive Wikipedia dump of 23 GB, containing articles spanning diverse topics and domains.
From this dataset, we randomly sampled 12.5k articles, each with a word count between 400 and 2000, for inclusion in our benchmark.

\subsection{Utilized LLMs and input prompts}\label{app:models}
\subsubsection{Models}
We employed seven distinct open-source large language models and their instruction-tuned variants sourced from the Hugging Face repository: \textbf{Llama3.2-1b}, \textbf{Llama3.2-3b}, \textbf{Llama3.1-8b}, \textbf{Mistral-7b}, \textbf{Qwen-7b}, \textbf{Gemma2-2b}, and \textbf{Gemma2-9b}.

\noindent
\textbf{Llama3.1-8b:} The Llama 3.1-8b-Instruct model is an instruction-tuned variant of Meta’s 8B-parameter LLM from the Llama 3.1 series, optimized for dialogue and assistant-like tasks. It is fine-tuned using supervised fine-tuning (SFT) and reinforcement learning with human feedback (RLHF), enhancing its ability to align with human preferences and generate helpful, and safe responses. The model supports up to 128k token context lengths and employs Grouped-Query Attention (GQA), rotary positional embeddings (RoPE), and SwiGLU activations to improve scalability and inference efficiency. Trained on over 15T tokens of publicly available data, it supports multilingual capabilities across several major languages and demonstrates strong performance in reasoning, text, and code generation tasks.

\noindent
\textbf{Llama3.2-1b} and \textbf{Llama3.2-3b:}
The Llama 3.2-1b-Instruct and Llama 3.2-3b-Instruct models are instruction-tuned variants of Meta's Llama 3.2 series, comprising 1.23 billion and 3.21 billion parameters respectively. Both models are optimized for multilingual dialogue applications, including tasks like agentic retrieval and summarization, and have demonstrated superior performance compared to many open-source and proprietary chat models on standard industry benchmarks. They employ an auto-regressive transformer architecture enhanced with GQA for improved inference scalability and support a context length of up to 128k tokens. Trained on up to 9 trillion tokens of publicly available online data, these models support multiple languages such as English, German, French, Italian, Portuguese, Hindi, Spanish, and Thai. 

\noindent
\textbf{Mistral-7b:}
We utilized Mistral-7b-Instruct-v0.3 model, that is an instruction-tuned variant of Mistral AI’s 7.25b-parameter language model, designed to excel in a wide range of natural language processing tasks. This version introduces several enhancements over its predecessors, including an expanded vocabulary of 32,768 tokens and support for the v3 tokenizer, which improves its ability to handle complex text inputs.
The v0.3 instruction-tuned variant has been specifically optimized through reinforcement learning from RLHF techniques to better follow user instructions and generate more helpful responses. Despite its relatively compact size compared to models with hundreds of billions of parameters, Mistral-7b-Instruct-v0.3 demonstrates competitive performance across various benchmarks, including reasoning, coding, and language understanding tasks.

\noindent
\textbf{Qwen-7b:}
The DeepSeek-R1-Distill-Qwen-7b model is a 7.62b-parameter instruction-tuned language model developed by DeepSeek AI, based on the Qwen2.5 architecture. It was fine-tuned using reinforcement learning and SFT techniques, leveraging reasoning data generated by the larger DeepSeek-R1 model. This training approach enhances the model's capabilities in reasoning, mathematics, and coding tasks. The model supports a context length of up to 128k tokens, enabling it to handle extensive inputs effectively.

\noindent
\textbf{Gemma2-2b} and \textbf{Gemma2-9b:}
The Gemma family, developed by Google, includes the Gemma2-2b-instruct and Gemma2-9b-instruct models, which are instruction-tuned variants of the base Gemma models. These models are part of Google's effort to provide lightweight, high-performing open models, drawing from the same research and technology used to create the Gemini models.
Both models are decoder-only large language models, available in English, and are designed to be versatile for various applications, including text generation, conversational AI, and summarization.

\begin{figure*}[t]
    \centering
    \begin{tabular}{cc}
     \small{Medium} & \small{News} \\
        \includegraphics[width=0.45\textwidth]{ 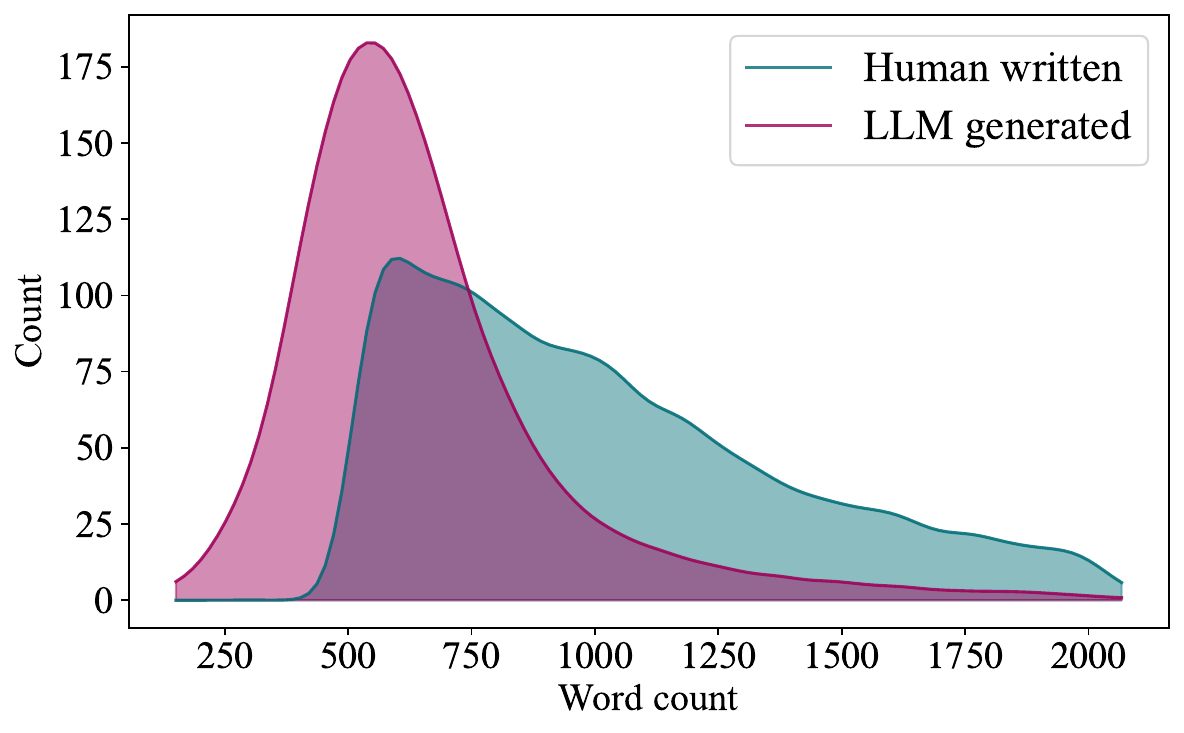} &
        \includegraphics[width=0.45\textwidth]{ 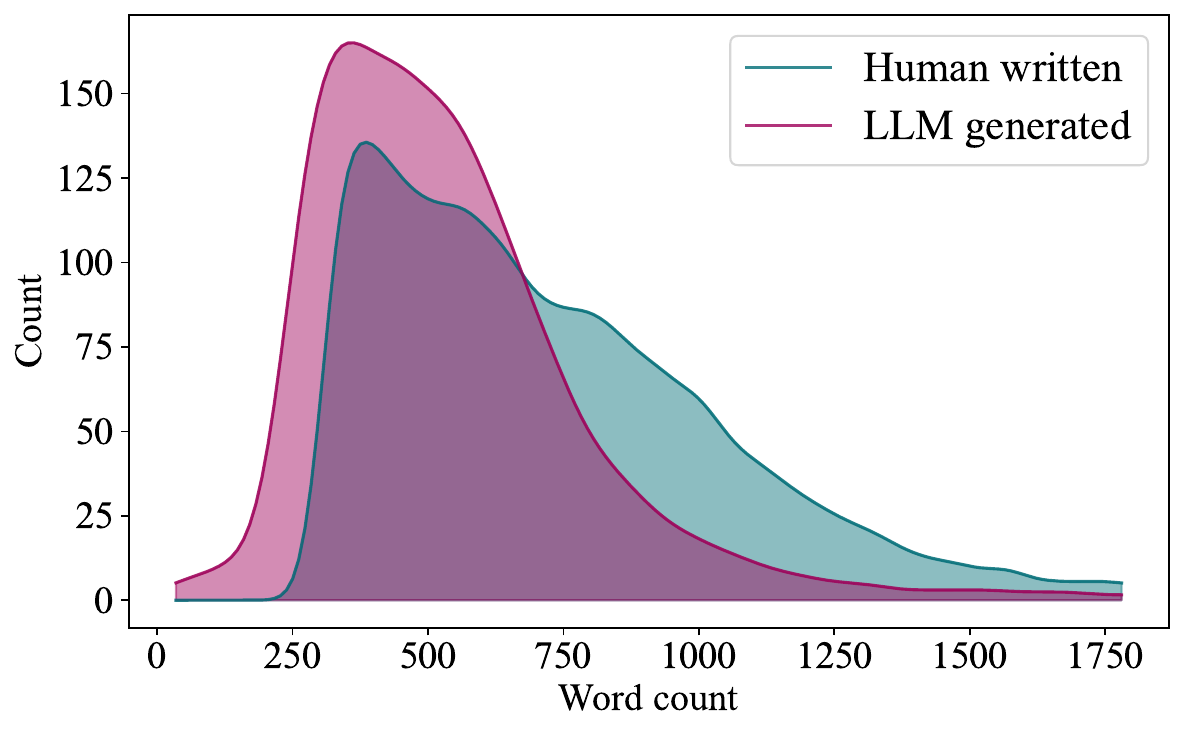} \\
        \small{Amazon reviews} &\small{Reddit} \\
        \includegraphics[width=0.45\textwidth]{ 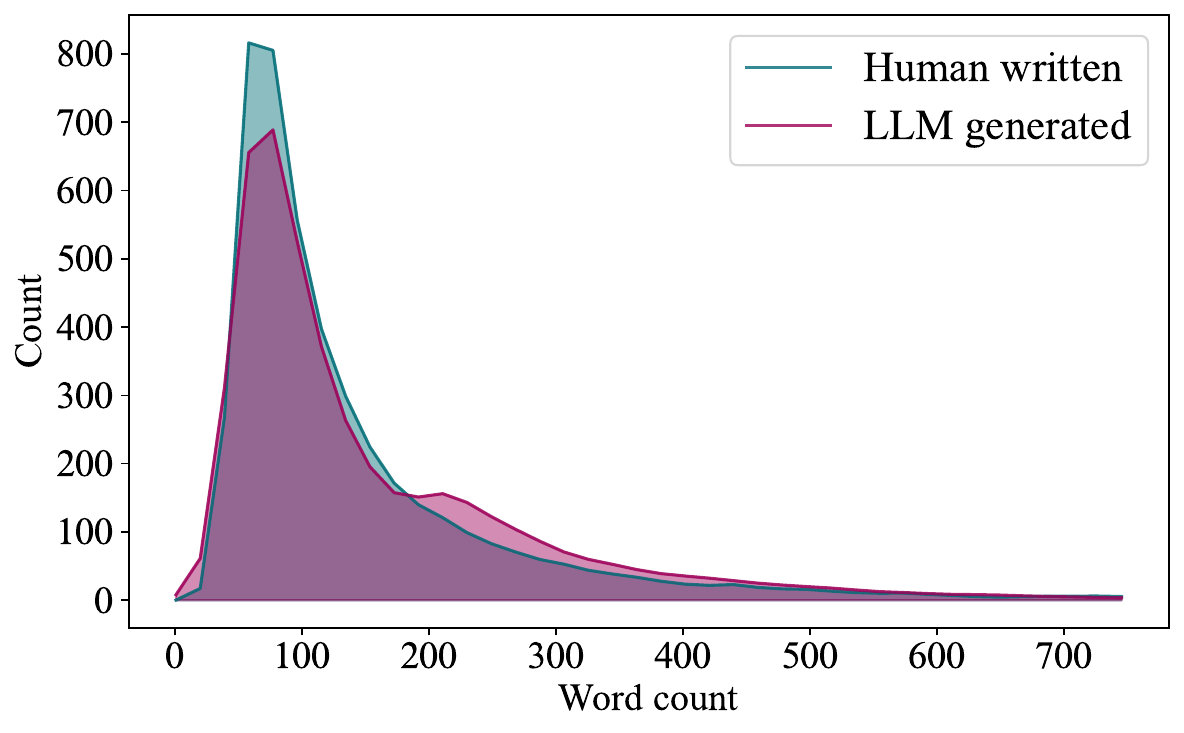} &
        \includegraphics[width=0.45\textwidth]{ 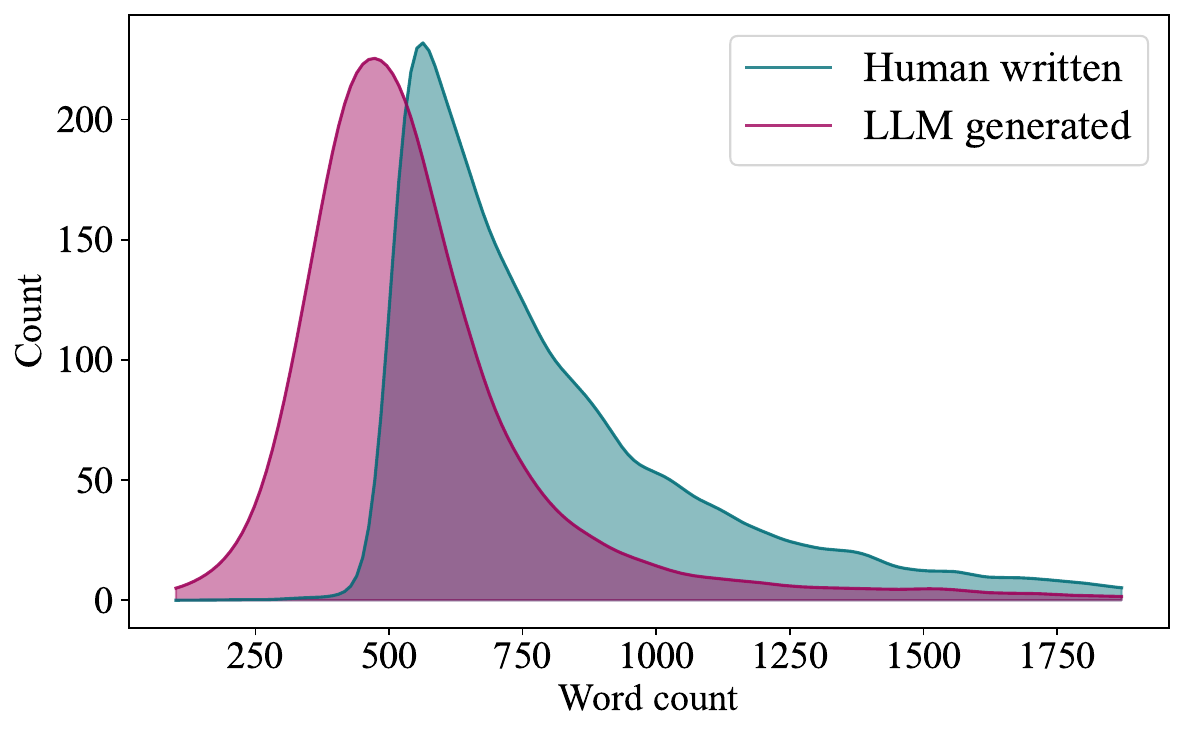} \\
         \small{arXiv} &\small{Pink slime} \\
        \includegraphics[width=0.45\textwidth]{ 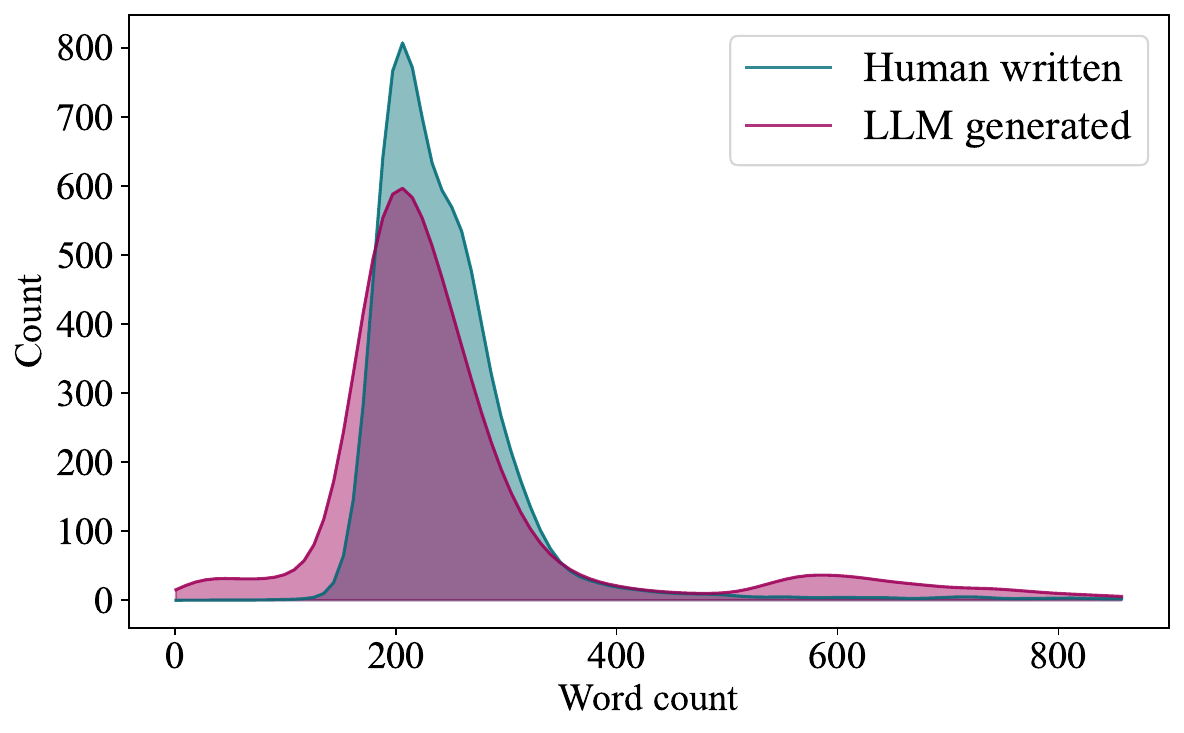} &
        \includegraphics[width=0.45\textwidth]{ 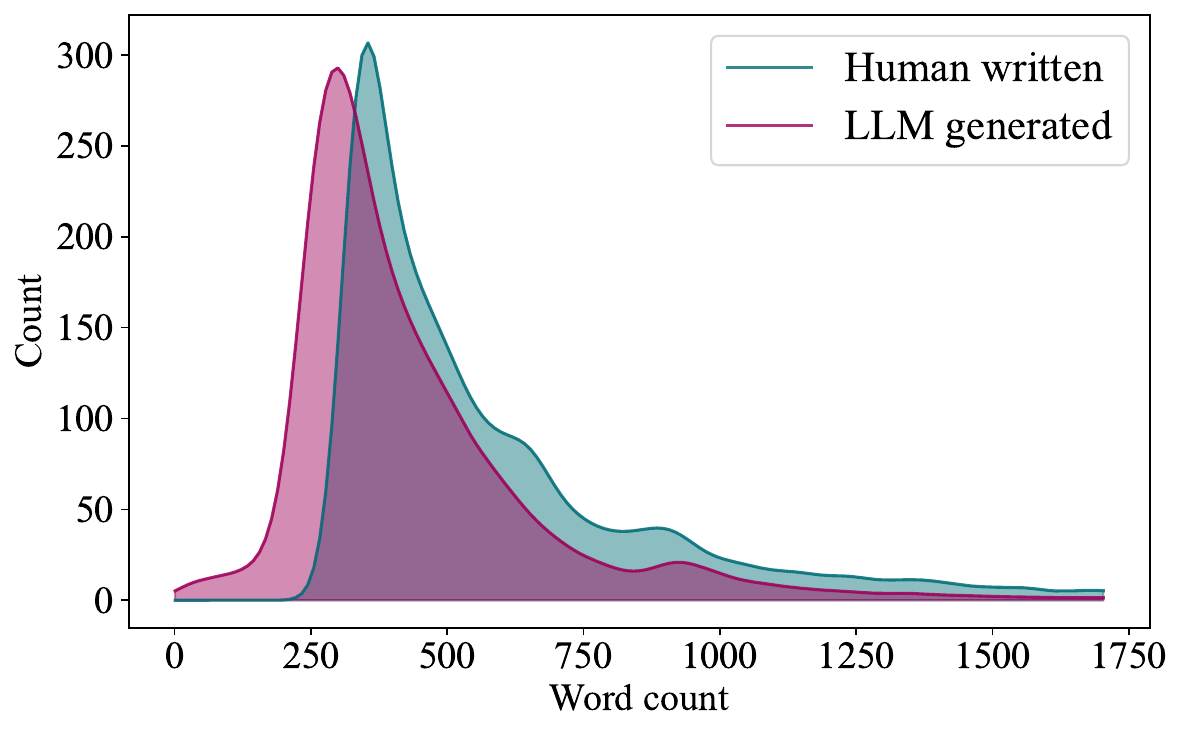} \\
        \small{Wikipedia} &\small{All} \\
        \includegraphics[width=0.45\textwidth]{ 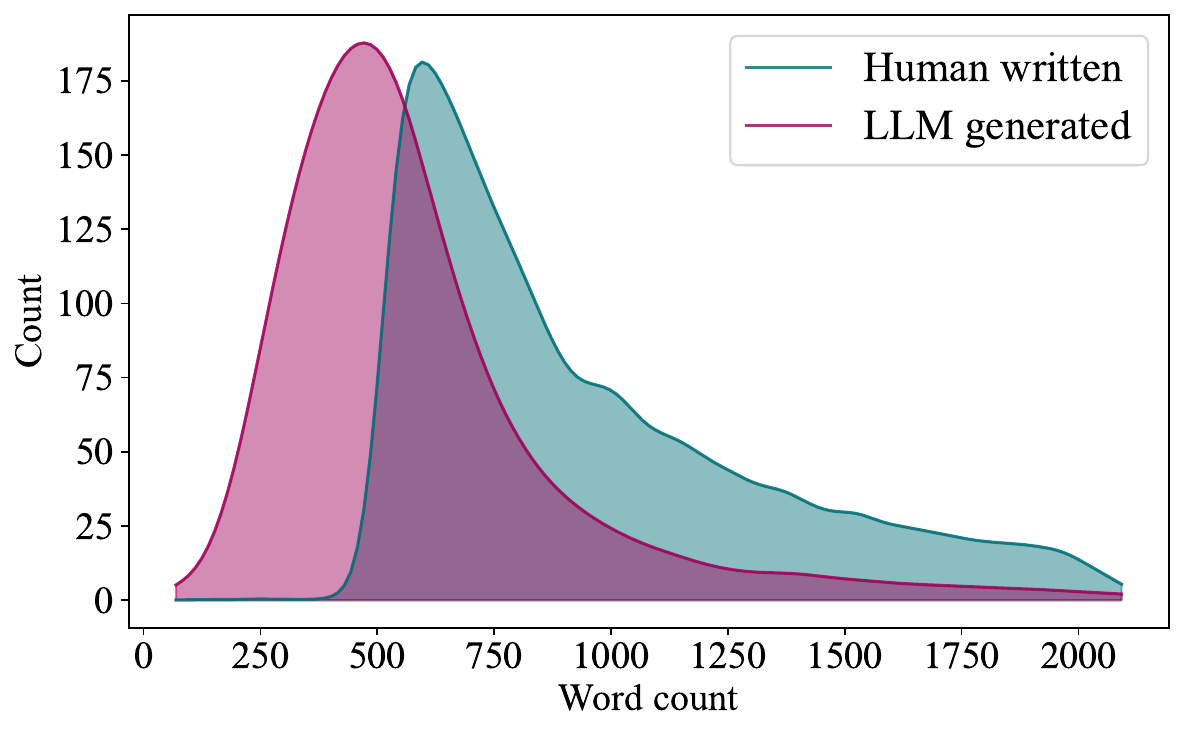} &
        \includegraphics[width=0.45\textwidth]{ 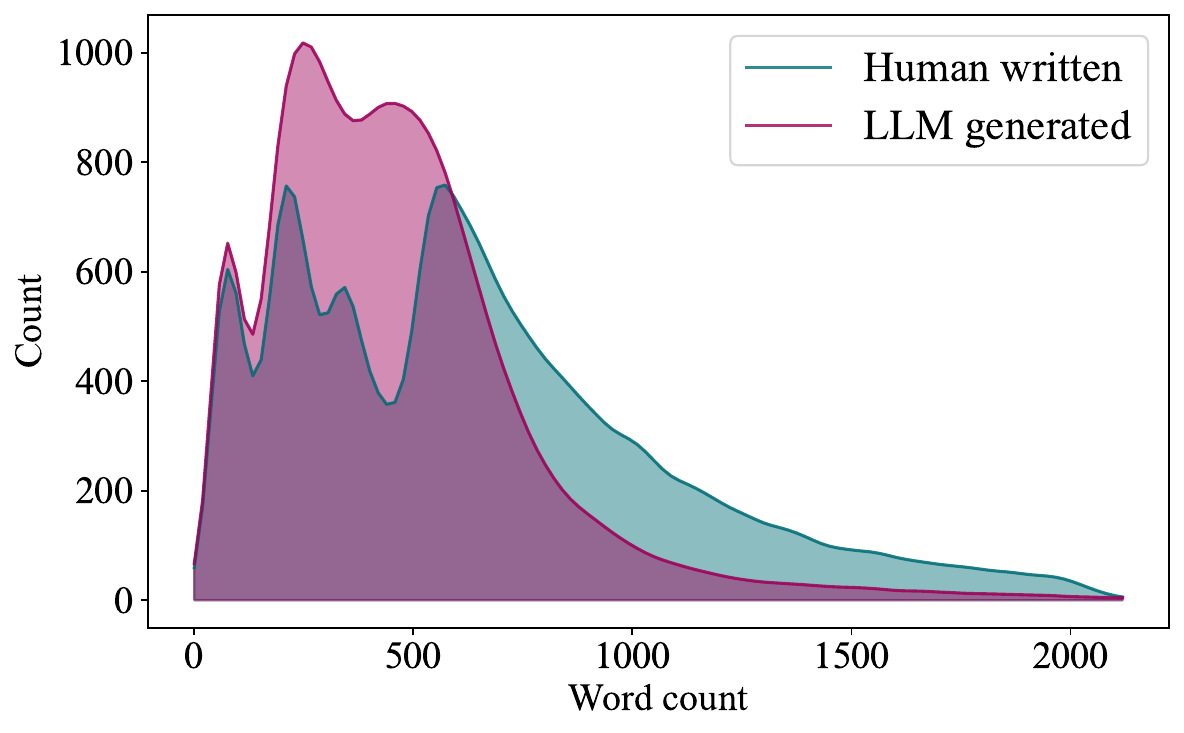}\\
       
    \end{tabular}
    \caption{Word count distribution of human-written and LLM-generated texts aggregated across all LLMs. ``All'' represents all samples across both writing styles and LLMs.}
    \label{fig:stats}
\end{figure*}
\subsubsection{Prompts}
We employed LLMs in their chat-based configurations to generate AI texts across multiple writing styles.
Each data domain received tailored prompts to elicit appropriate responses.
For the ``pink slime'' data, we supplemented prompts with definitional context to ensure semantic alignment. 
To preserve comparability in length, the LLMs were instructed to produce outputs approximately equal in word count to the corresponding human-written text (denoted as \texttt{<N>}).
Moreover, in the Amazon Reviews, Reddit, Pink slime, and Wikipedia datasets, we incorporated the respective titles, product name, post title, news headline, or document title, into the input prompt to enhance coherence and topical relevance.
Below, we detail the specific prompts used to condition LLM outputs across the various writing styles examined in this work.

\noindent
\textbf{Medium}:
\begin{tcolorbox}[colback=gray!10, colframe=black!50, boxrule=0.4pt, arc=2pt, left=1mm, right=1mm, top=1mm, bottom=1mm]
\ttfamily
[\{`role': `system', `content': `You are a blog writer in Medium website. You paraphrase the Medium article I give you in about <N> words as if you are the original author, maintaining the same ideas and tone while using your own words.'\},\\\{`role': `user', `content': `The article is <HUMAN TEXT>'\},]
\end{tcolorbox}
\noindent
\textbf{News}:
\begin{tcolorbox}[colback=gray!10, colframe=black!50, boxrule=0.4pt, arc=2pt, left=1mm, right=1mm, top=1mm, bottom=1mm]
\ttfamily
[\{`role': `system', `content': `You are a journalist working for a reputable news agency. You paraphrase the news article I give you in about <N> words as if you are the original writer, maintaining the same ideas and tone while using your own words.'\},\\\{`role': `user', `content': `The article is <HUMAN TEXT>''\},]
\end{tcolorbox}
\noindent
\textbf{Amazon reviews}:
\begin{tcolorbox}[colback=gray!10, colframe=black!50, boxrule=0.4pt, arc=2pt, left=1mm, right=1mm, top=1mm, bottom=1mm]
\ttfamily
[\{`role': `system', `content': `You are an Amazon customer. You paraphrase the review I give you about a product with title <TITLE> in about <N> words as if you are the original review writer, maintaining the same ideas and tone while using your own words.'\},\\\{`role': `user', `content': `The article is <HUMAN TEXT>''\},]
\end{tcolorbox}
\noindent
\textbf{Reddit}:
\begin{tcolorbox}[colback=gray!10, colframe=black!50, boxrule=0.4pt, arc=2pt, left=1mm, right=1mm, top=1mm, bottom=1mm]
\ttfamily
[\{`role': `system', `content': `You are a Reddit user. You paraphrase the Reddit post with title <TITLE> I give you in about <N> words as if you are the original Reddit post writer, maintaining the same ideas and tone while using your own words.'\},\\\{`role': `user', `content': `The article is <HUMAN TEXT>''\},]
\end{tcolorbox}
\noindent
\textbf{arXiv}:
\begin{tcolorbox}[colback=gray!10, colframe=black!50, boxrule=0.4pt, arc=2pt, left=1mm, right=1mm, top=1mm, bottom=1mm]
\ttfamily
[\{`role': `system', `content': `You are a scientific paper writer. You paraphrase the abstract of a scientific paper I give you in about <N> words as if you are the original author, maintaining the same ideas and tone while using your own words.'\},\\\{`role': `user', `content': `The article is <HUMAN TEXT>''\},]
\end{tcolorbox}
\noindent
\textbf{Pink slime}:
\begin{tcolorbox}[colback=gray!10, colframe=black!50, boxrule=0.4pt, arc=2pt, left=1mm, right=1mm, top=1mm, bottom=1mm]
\ttfamily
[\{`role': `system', `content': `Pink slime journalism is a practice in which American news outlets, or fake partisan operations masquerading as such, publish poor-quality news reports which appear to be local news. You are a pink slime journalist. You paraphrase the pink slime article I give you about a subject with title <TITLE> in about <N> words as if you are the original article writer, maintaining the same ideas and tone while using your own words.'\},\\\{`role': `user', `content': `The article is <HUMAN TEXT>''\},]
\end{tcolorbox}
\noindent
\textbf{Wikipedia}:
\begin{tcolorbox}[colback=gray!10, colframe=black!50, boxrule=0.4pt, arc=2pt, left=1mm, right=1mm, top=1mm, bottom=1mm]
\ttfamily
[\{`role': `system', `content': `You are a Wikipedia writer. You paraphrase the article I give you about a subject with title <TITLE> in about <N> words as if you are the original article writer, maintaining the same ideas and tone while using your own words.'\},\\\{`role': `user', `content': `The article is <HUMAN TEXT>''\},]
\end{tcolorbox}

\subsection{Dataset statistics}\label{app:stat}

\begin{table*}[t]
\centering
\caption{Data statistics for each LLM utilized.}
\begin{adjustbox}{max width=\textwidth}
\begin{tabular}{l|ccc|ccc|ccc|ccc}
\toprule
\textbf{Dataset $\rightarrow$} & \multicolumn{3}{c|}{\textbf{Medium}} & \multicolumn{3}{c|}{\textbf{News}} & \multicolumn{3}{c|}{\textbf{Amazon reviews}}& \multicolumn{3}{c}{\textbf{Reddit}}\\
\textbf{Strategy $\downarrow$} & Human & Paraphrased & Humanified &  Human & Paraphrased & Humanified&  Human & Paraphrased & Humanified &  Human & Paraphrased & Humanified \\
\bottomrule
Paraphrasing & 3000 &3000 & \textbf{---}&3000 &3000 &\textbf{---}&3000 &3000 &\textbf{---}&3000 &3000 &\textbf{---} \\
\midrule
RMM @ $p{=}10\%$ &500&500&500&500&500&500&500&500&500&500&500&500 \\
RMM @ $p{=}20\%$&500&500&500&500&500&500&500&500&500&500&500&500 \\
RMM @ $p{=}40\%$&500&500&500&500&500&500&500&500&500&500&500&500 \\
RMM @ $p{=}60\%$&500&500&500&500&500&500&500&500&500&500&500&500 \\
RMM @ $p{=}80\%$&500&500&500&500&500&500&500&500&500&500&500&500 \\
RMM @ $p{=}100\%$&500&500&500&500&500&500&500&500&500&500&500&500 \\
\midrule
AWS @ $p{=}10\%$ &500&500&500&500&500&500&500&500&500&500&500&500 \\
AWS @ $p{=}20\%$&500&500&500&500&500&500&500&500&500&500&500&500 \\
AWS @ $p{=}40\%$&500&500&500&500&500&500&500&500&500&500&500&500 \\
AWS @ $p{=}60\%$&500&500&500&500&500&500&500&500&500&500&500&500 \\
AWS @ $p{=}80\%$&500&500&500&500&500&500&500&500&500&500&500&500 \\
AWS @ $p{=}100\%$&500&500&500&500&500&500&500&500&500&500&500&500 \\
\midrule
RHL @ $R{=}5\%$ &500&500&500&500&500&500&500&500&500&500&500&500 \\
RHL @ $R{=}15\%$&500&500&500&500&500&500&500&500&500&500&500&500 \\
RHL @ $R{=}20\%$&500&500&500&500&500&500&500&500&500&500&500&500 \\
RHL @ $R{=}25\%$&500&500&500&500&500&500&500&500&500&500&500&500 \\
RHL @ $R{=}30\%$&500&500&500&500&500&500&500&500&500&500&500&500 \\
RHL @ $R{=}35\%$&500&500&500&500&500&500&500&500&500&500&500&500 \\
RHL @ $R{=}40\%$&500&500&500&500&500&500&500&500&500&500&500&500 \\
\bottomrule
\bottomrule
\textbf{Dataset $\rightarrow$} & \multicolumn{3}{c|}{\textbf{arXiv}} & \multicolumn{3}{c|}{\textbf{Pink slime}} & \multicolumn{3}{c|}{\textbf{Wikipedia}}&  \\
\textbf{Strategy $\downarrow$} & Human & Paraphrased & Humanified &  Human & Paraphrased & Humanified&  Human & Paraphrased & Humanified &    \\
\bottomrule
Paraphrasing & 3000 &3000 & \textbf{---}&3000 &3000 &\textbf{---}&3000 &3000 &\textbf{---} \\
\midrule
RMM @ $p{=}10\%$ &500&500&500&500&500&500&500&500&50 \\
RMM @ $p{=}20\%$&500&500&500&500&500&500&500&500&500\\
RMM @ $p{=}40\%$&500&500&500&500&500&500&500&500&500\\
RMM @ $p{=}60\%$&500&500&500&500&500&500&500&500&500\\
RMM @ $p{=}80\%$&500&500&500&500&500&500&500&500&500\\
RMM @ $p{=}100\%$&500&500&500&500&500&500&500&500&50 \\
\midrule
AWS @ $p{=}10\%$ &500&500&500&500&500&500&500&500&50 \\
AWS @ $p{=}20\%$&500&500&500&500&500&500&500&500&500\\
AWS @ $p{=}40\%$&500&500&500&500&500&500&500&500&500\\
AWS @ $p{=}60\%$&500&500&500&500&500&500&500&500&500\\
AWS @ $p{=}80\%$&500&500&500&500&500&500&500&500&500\\
AWS @ $p{=}100\%$&500&500&500&500&500&500&500&500&50 \\
\midrule
RHL @ $R{=}5\%$ &500&500&500&500&500&500&500&500&500\\
RHL @ $R{=}15\%$&500&500&500&500&500&500&500&500&500\\
RHL @ $R{=}20\%$&500&500&500&500&500&500&500&500&500\\
RHL @ $R{=}25\%$&500&500&500&500&500&500&500&500&500\\
RHL @ $R{=}30\%$&500&500&500&500&500&500&500&500&500\\
RHL @ $R{=}35\%$&500&500&500&500&500&500&500&500&500\\
RHL @ $R{=}40\%$&500&500&500&500&500&500&500&500&500\\
\bottomrule
\end{tabular}
\end{adjustbox}
\label{tab:stats}
\end{table*}
Table \ref{tab:stats} reports the number of text samples included in our benchmark for each LLM evaluated.
The ``Human'' column denotes the number of collected human-written texts.
The ``Paraphrased'' column corresponds to the original LLM-generated outputs without humanification.
The ``Humanified'' column indicates the number of paraphrased samples that were modified using the humanification strategies specified in the corresponding rows.
Additionally, Figure \ref{fig:stats} illustrates the word count distributions of human-written texts and LLM-generated paraphrased outputs for each writing style, aggregated across all LLMs used in this study.
\section{Text samples}\label{app:samples}
For masked word prediction within our MLM framework, we employed the Longformer-base-4096 architecture developed by the Allen Institute for AI \cite{Beltagy2020Longformer}.
In this section, we provide exemplar texts demonstrating the three humanification strategies employed in our methodology.
The words enclosed in brackets and highlighted in red represent words predicted by the MLM architecture, which subsequently replaced their corresponding precedent words in the paraphrased text.
Text highlighted in blue represents the baseline AI-paraphrased content prior to humanification processing.
\subsection{Sample humanified text from RMM strategy}

{\color{blue}
The White House is escalating its efforts to persuade Congress to approve limited \textcolor{red}{[military]} strikes \textcolor{red}{[action]} against Syria, as President Obama faces a formidable challenge in convincing lawmakers to back a new military campaign in the Middle East. The president has been personally engaging with skeptical lawmakers over the weekend, delivering \textcolor{red}{[making]} a tailored pitch to Democrats and Republicans who remain undecided or open \textcolor{red}{[close]} to reconsidering their opposition to military action.

According to White House officials, Obama's argument \textcolor{red}{[case]} centers on the dual imperative of both moral responsibility and national security \textcolor{red}{interest}. The president believes that the United States has a critical obligation to respond to the devastating chemical weapons attack \textcolor{red}{[attacks]} in Syria, which has left countless civilians, including children, dead or injured. This perspective is underscored by a series of disturbing videos, obtained by ABC News, which were shown to lawmakers in a classified briefing \textcolor{red}{[session]} last week.

These graphic images, which depict the harrowing aftermath of the chemical attack \textcolor{red}{[attacks]}, are being used by the administration to make a powerful \textcolor{red}{[compelling]} case to Congress and the American public. Secretary of State John Kerry, who has been leading the charge to build international support \textcolor{red}{[consensus]} for military action, referenced the videos in a speech in Paris on Saturday, emphasizing the atrocities committed by the Syrian \textcolor{red}{[Assad]} regime against its own people \textcolor{red}{[citizens]}.

Kerry's impassioned plea, which highlighted the tragic fate of innocent civilians, including children, is a stark reminder \textcolor{red}{[illustration]} of the human cost of inaction. As he noted, the use of chemical weapons in the middle of the night, when people \textcolor{red}{[children]} should have been sleeping safely in their beds, is an unconscionable act that demands \textcolor{red}{[warrants]} a response from the international community \textcolor{red}{[body]}.

The administration is aware that it faces a tough sell in convincing Congress to approve military action, with a recent ABC News survey indicating deep opposition among lawmakers. However, officials remain hopeful that they can build a coalition of 60 \textcolor{red}{[enough]} Democrats and Republicans to overcome the threat of a filibuster and secure approval for the military strike.

In a bid to build support \textcolor{red}{[consensus]}, Vice President Biden is hosting a dinner for over a dozen Republican senators on Sunday night, with the guest list including several key lawmakers who have expressed reservations about military action. The administration is also counting on the support \textcolor{red}{[endorsement]} of retired General David Petraeus, a respected figure in military circles, who has publicly urged lawmakers to back the military strike.

As the vote on the military strike looms in the Senate as early as Wednesday, the White House is stepping \textcolor{red}{[speeding]} up its efforts to persuade lawmakers to support \textcolor{red}{[back]} the president's plan. Obama will make his case to the American public in a series of interviews with network television anchors \textcolor{red}{[stations]} on Monday, followed by a televised address on Tuesday. The outcome of this high-stakes debate will have far-reaching \textcolor{red}{[wide-reaching]} implications for the United States and its role in the Middle East, and the White House is leaving \textcolor{red}{[left]} no stone unturned in its bid to secure approval for military action.
}
\subsection{Sample humanified text from AWS strategy}
{\color{blue}
Pleased with the affordable \textcolor{red}{[good]} price and adorable \textcolor{red}{[good]} design[ \textcolor{red}{[look]}, I've been using this cardigan for work over the past few months. However \textcolor{red}{[but]}, I've noticed minor \textcolor{red}{[some]} issues \textcolor{red}{[things]} where my arms have caused small fabric balls to form, which can be a bit bothersome but not entirely unexpected \textcolor{red}{[bad]} for a budget-friendly \textcolor{red}{[work-type]} item. Sizing has been an issue \textcolor{red}{[problem]} for me, as I often \textcolor{red}{[have]} find \textcolor{red}{[that]} cardigans in my size still gape open. If you're larger-chested \textcolor{red}{[big-chested]} like me, I'd suggest sizing up for a better fit. Despite \textcolor{red}{[Besides]} these concerns \textcolor{red}{[things]}, I might \textcolor{red}{[have]} still purchase \textcolor{red}{[get]} more due \textcolor{red}{[because]} to the low cost, but a higher-quality \textcolor{red}{[higher-cut]} garment might \textcolor{red}{[do]} offer \textcolor{red}{[get]} better longevity \textcolor{red}{[looks]}.
}

\subsection{Sample humanified text from RHL strategy}
{\color{blue}
I'm feeling \textcolor{red}{[really]} so hurt \textcolor{red}{[upset]} and confused after our last conversation.
I said I was sorry and wanted to make amends \textcolor{red}{[up]}, but he's now acting like I'm the one who's done him wrong. I don't see how I deserve \textcolor{red}{[want]} to be treated \textcolor{red}{[loved]} this way, especially \textcolor{red}{[just]} when I've been trying \textcolor{red}{[able]} to be understanding \textcolor{red}{[kind]} and accommodating \textcolor{red}{[loving]}. Our relationship was pretty \textcolor{red}{[so]} intense \textcolor{red}{[good]} from the start, and we were monogamous pretty \textcolor{red}{[very]} quickly. However \textcolor{red}{[unfortunately]}, he refused \textcolor{red}{[wanted]} to acknowledge \textcolor{red}{[believe]} that we were in a real relationship for a few months, which was confusing and frustrating \textcolor{red}{[annoying]}. He had high expectations \textcolor{red}{[opinion]} of me, and I felt like I had to constantly prove \textcolor{red}{[explain]} myself to him. I've never \textcolor{red}{[only]} cheated on anyone before, and I thought I was able to handle \textcolor{red}{[explain]} myself around \textcolor{red}{[in]} people with good intentions \textcolor{red}{[reason]}. One of the things that really bothered me was how he would belittle \textcolor{red}{[tell]} me about my male friends. He would get \textcolor{red}{[be]} upset if I spent time with them or made plans with them, and he would even get \textcolor{red}{[be]} angry if I didn't include them in everything. I felt like he was conditioning me into isolation, and it was suffocating \textcolor{red}{[annoying]}. There were a lot \textcolor{red}{[couple]} of fights in our relationship, and many of them were petty \textcolor{red}{[stupid]}. Like the time he thought I smelled \textcolor{red}{[was]} of cologne \textcolor{red}{[stupid]} and swore \textcolor{red}{[that]} I had just slept with someone \textcolor{red}{[her]}. Or the time he came over and claimed \textcolor{red}{[said]} there were tire marks \textcolor{red}{[nuts]} in his parking \textcolor{red}{[sweet]} spot, even though \textcolor{red}{[if]} it was just a dirt \textcolor{red}{[back]} road. It was like he was looking for any excuse to get \textcolor{red}{[be]} upset. I also \textcolor{red}{[really]} felt like he was playing games with me. Like the time I told him I was lazy \textcolor{red}{[sick]} and didn't want to drive \textcolor{red}{[go]} to my friends' house, and he got upset because I didn't spend the night \textcolor{red}{[time]} with him. Or the time he broke up with me and then said we weren't in a relationship, even though \textcolor{red}{[if]} we had been hanging out and talking. It was like he was trying \textcolor{red}{[able]} to manipulate \textcolor{red}{[trick]} me into feeling \textcolor{red}{[being]} guilty \textcolor{red}{[stupid]} or responsible \textcolor{red}{[sorry]} for his own emotions \textcolor{red}{[faults]}. But the thing that really hurt \textcolor{red}{[upset]} me was when he introduced me to his ex-girlfriend \textcolor{red}{[girl-friends]} as his friend, and then acted like I was the one who had done him wrong. He said I was "fucking with his heart and emotions \textcolor{red}{[mind]}," and that I needed to earn \textcolor{red}{[lose]} his trust. It was like he was trying \textcolor{red}{[out]} to make me feel \textcolor{red}{[look]} like I was the problem \textcolor{red}{[one]}, even though \textcolor{red}{[if]} I had done nothing wrong. I feel \textcolor{red}{[felt]} like I'm losing \textcolor{red}{[wasting]} my mind, to be honest \textcolor{red}{[fair]}. I'm trying \textcolor{red}{[attempting]} to be understanding \textcolor{red}{[kind]} and accommodating \textcolor{red}{[sweet]}, but it feels \textcolor{red}{[was]} like he's not giving me any space or trust. I'm starting \textcolor{red}{[beginning]} to wonder \textcolor{red}{[think]} if I'm just not good enough for him, or if he's just not willing \textcolor{red}{[able]} to work through our issues \textcolor{red}{[shit]} together. Do I deserve \textcolor{red}{[have]} to be treated \textcolor{red}{[left]} this way?
}
\section{Detectors} \label{app:detectors}
\noindent
\textbf{Binoculars} \cite{hans2024spotting} is a zero-shot, model-agnostic method for detecting machine-generated text that requires no training data. It operates by comparing the perplexity of a text as evaluated by two language models: an ``observer'' model and a ``performer'' model. The observer computes the perplexity of the text directly, while the performer generates next-token predictions, which are then evaluated by the observer to compute cross-perplexity. The ratio of perplexity to cross-perplexity serves as a strong indicator of whether the text is human- or machine-generated.

\noindent
\textbf{Fast-DetectGPT} \cite{bao2023fast} is a zero-shot method, building upon the principles of DetectGPT. It introduces the concept of conditional probability curvature to distinguish between human- and AI-authored content. The method operates by sampling alternative word choices for a given text and evaluating the conditional probabilities using a language model. By analyzing the curvature of these probabilities, Fast-DetectGPT identifies AI text. 

\noindent
\textbf{LRR} (Log-Likelihood Log-Rank Ratio) \cite{su2023detectllm} is a zero-shot approach. It combines two statistical measures: the log-likelihood, which assesses the absolute confidence of a language model in predicting a sequence, and the log-rank, which evaluates the relative ranking of the predicted tokens. By computing the ratio of these two measures, LRR captures nuanced differences between human-written and LLM-generated text. 

\noindent
\textbf{Log-Likelihood} \cite{gehrmann-etal-2019-gltr} calculates the log-probability of each token in a text sequence using a language model, assessing how predictable each word is within its context. In this framework, human-written text typically exhibits a mix of high- and low-probability tokens, reflecting natural linguistic variability. In contrast, LLM-generated text often contains a higher proportion of high-probability tokens, indicating more predictable word choices.

\noindent
\textbf{Rank} \cite{gehrmann-etal-2019-gltr} evaluates the predictability of each token in a text by determining its rank within the language model's probability distribution. Tokens that consistently appear among the top-ranked predictions indicate higher predictability, a characteristic often associated with LLM-generated text. Analyzing the distribution of these token ranks assists in distinguishing between human-authored and AI-generated content. 

\noindent
\textbf{Log-Rank} \cite{gehrmann-etal-2019-gltr} enhances the performance of Rank method by applying a logarithmic transformation to the rank of each token within a language model's predicted probability distribution. 

\noindent
\textbf{RADAR} \cite{NEURIPS2023_30e15e59} is a framework designed to enhance the detection of AI-generated text, particularly against paraphrased content that often evades traditional detectors. It employs an adversarial training approach involving two components: a paraphraser and a detector. The paraphraser aims to rewrite AI-generated text to resemble human-authored content, thereby challenging the detector's ability to identify machine-generated text. Conversely, the detector is trained to distinguish between human-written and AI-generated texts.

\section{Additional results}\label{app:smallmodel}
\begin{figure*}[t]
    \centering
    \setlength{\tabcolsep}{2pt}
    \renewcommand{\arraystretch}{1}
    \begin{tabular}{cccc}
     \small{Llama3.2 1b} &  \small{Llama3.2 3b} & \small{Llama3.1 8b} & \small{Mistral 7b}\\
        \includegraphics[width=0.2\textwidth]{ 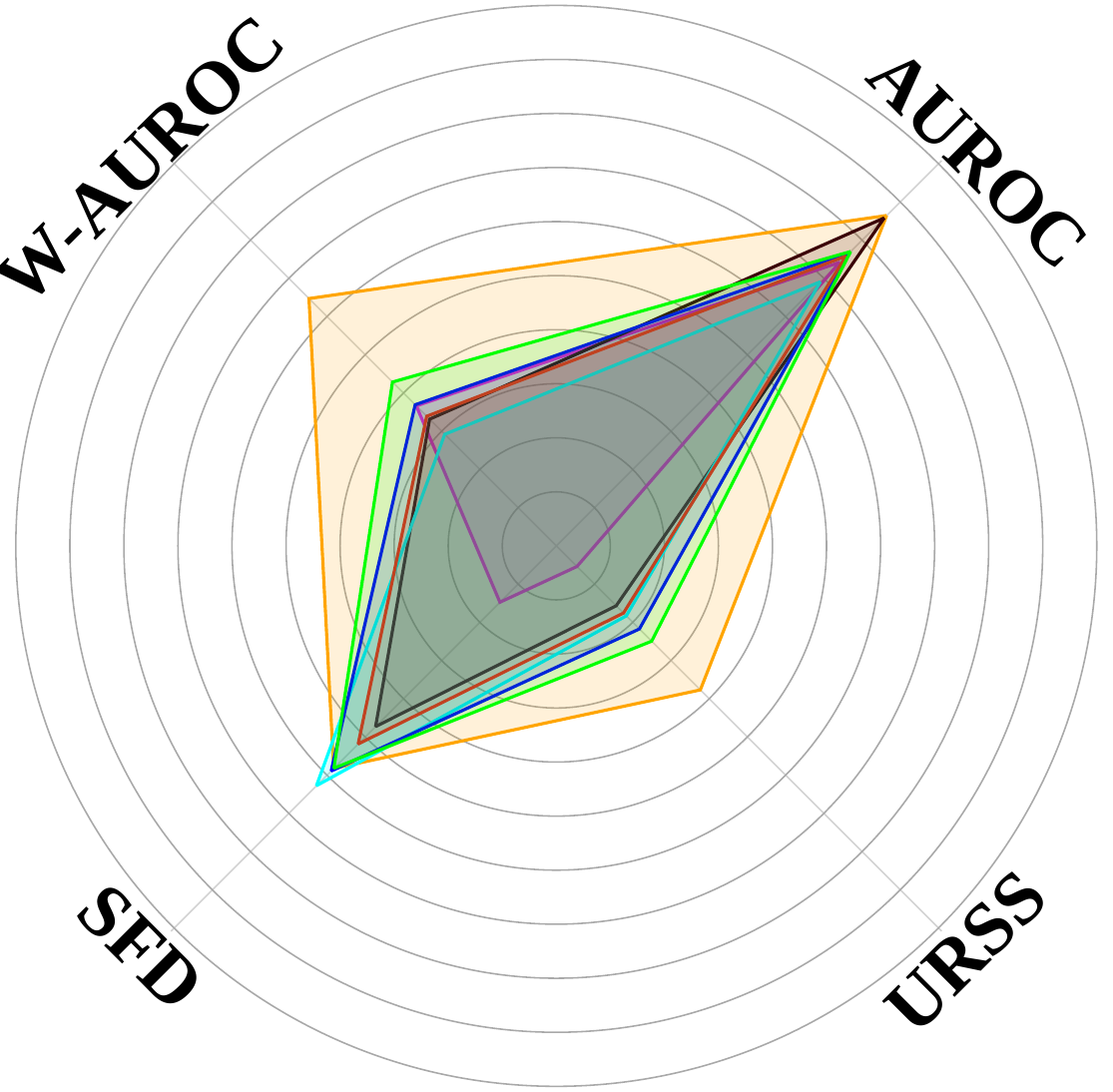} &
        \includegraphics[width=0.2\textwidth]{ 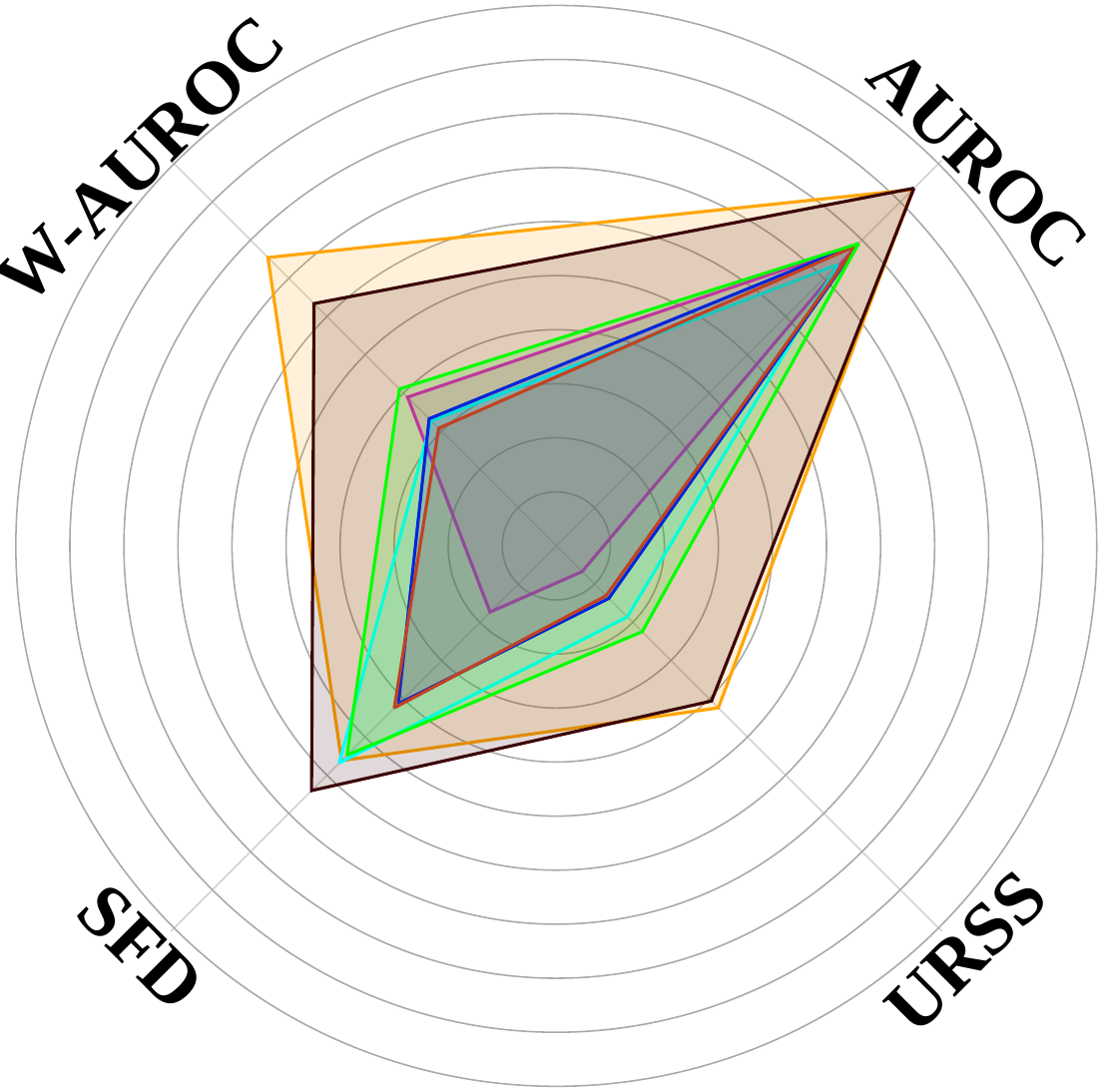} &
        \includegraphics[width=0.2\textwidth]{ 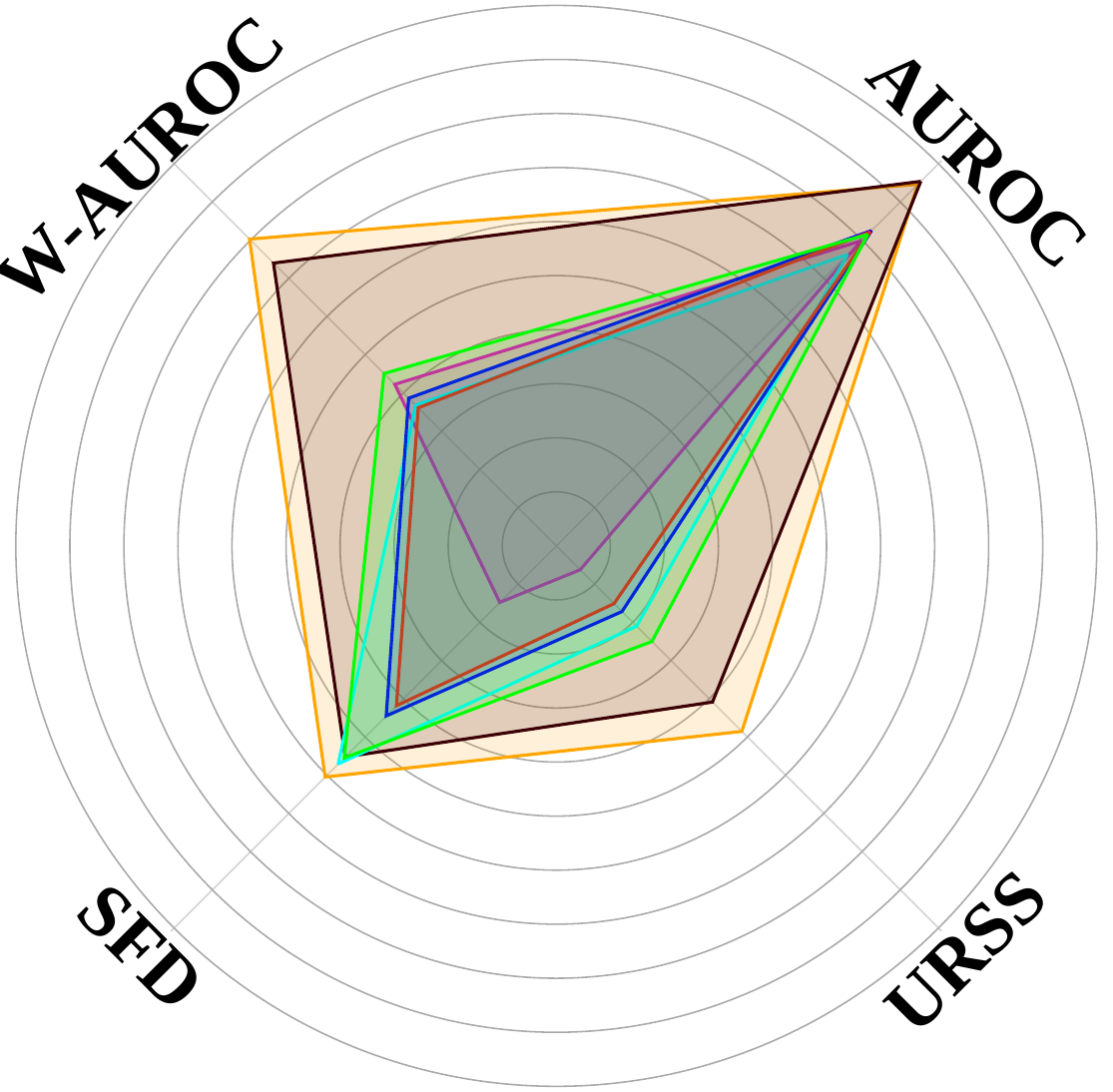} &
        \includegraphics[width=0.2\textwidth]{ 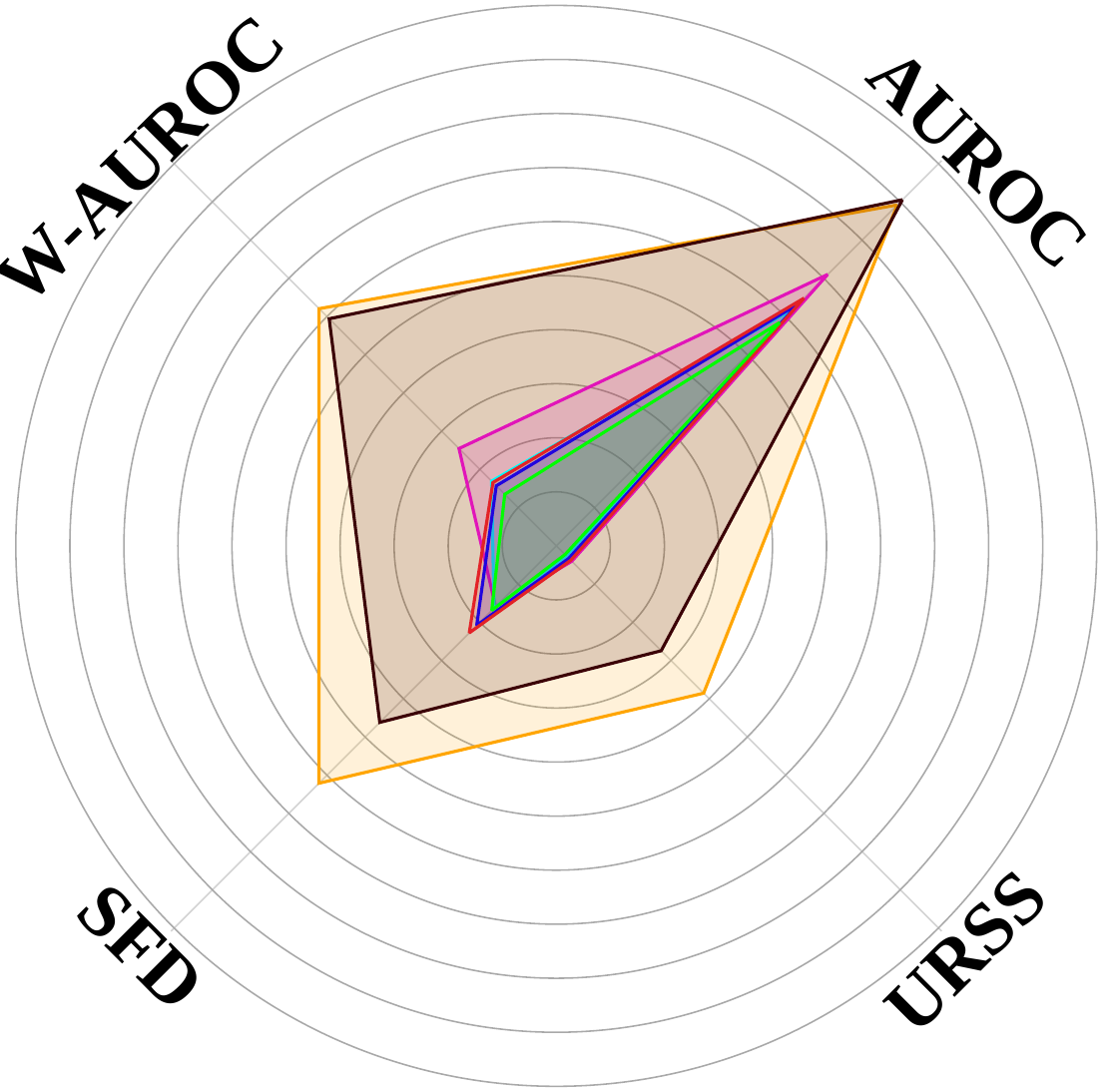} \\
       \small{Qwen 7b} & \small{Gemma2 2b} & \small{Gemma2 9b} &  \\
        \includegraphics[width=0.2\textwidth]{ 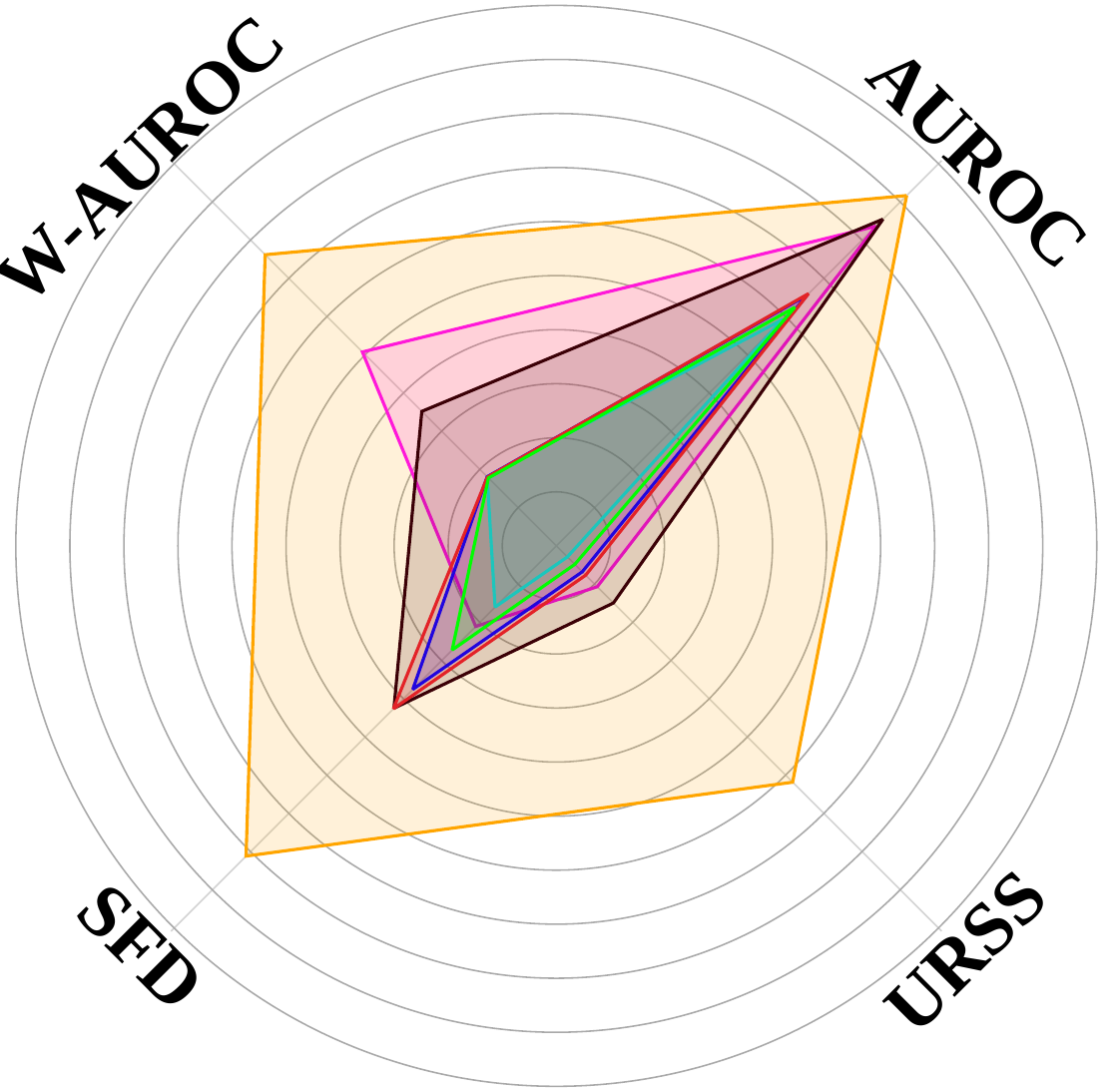} &
        \includegraphics[width=0.2\textwidth]{ 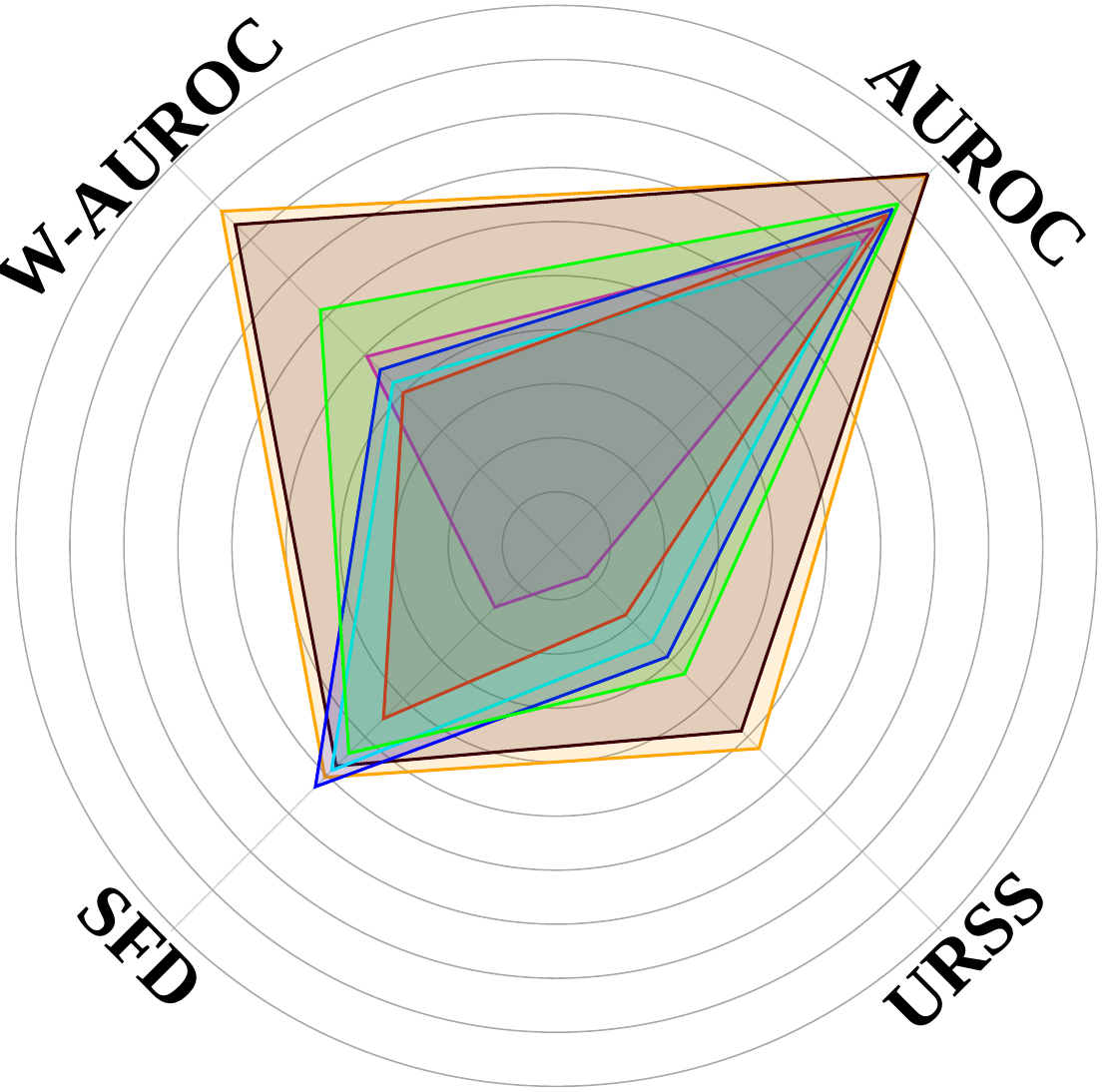} &
        \includegraphics[width=0.2\textwidth]{ 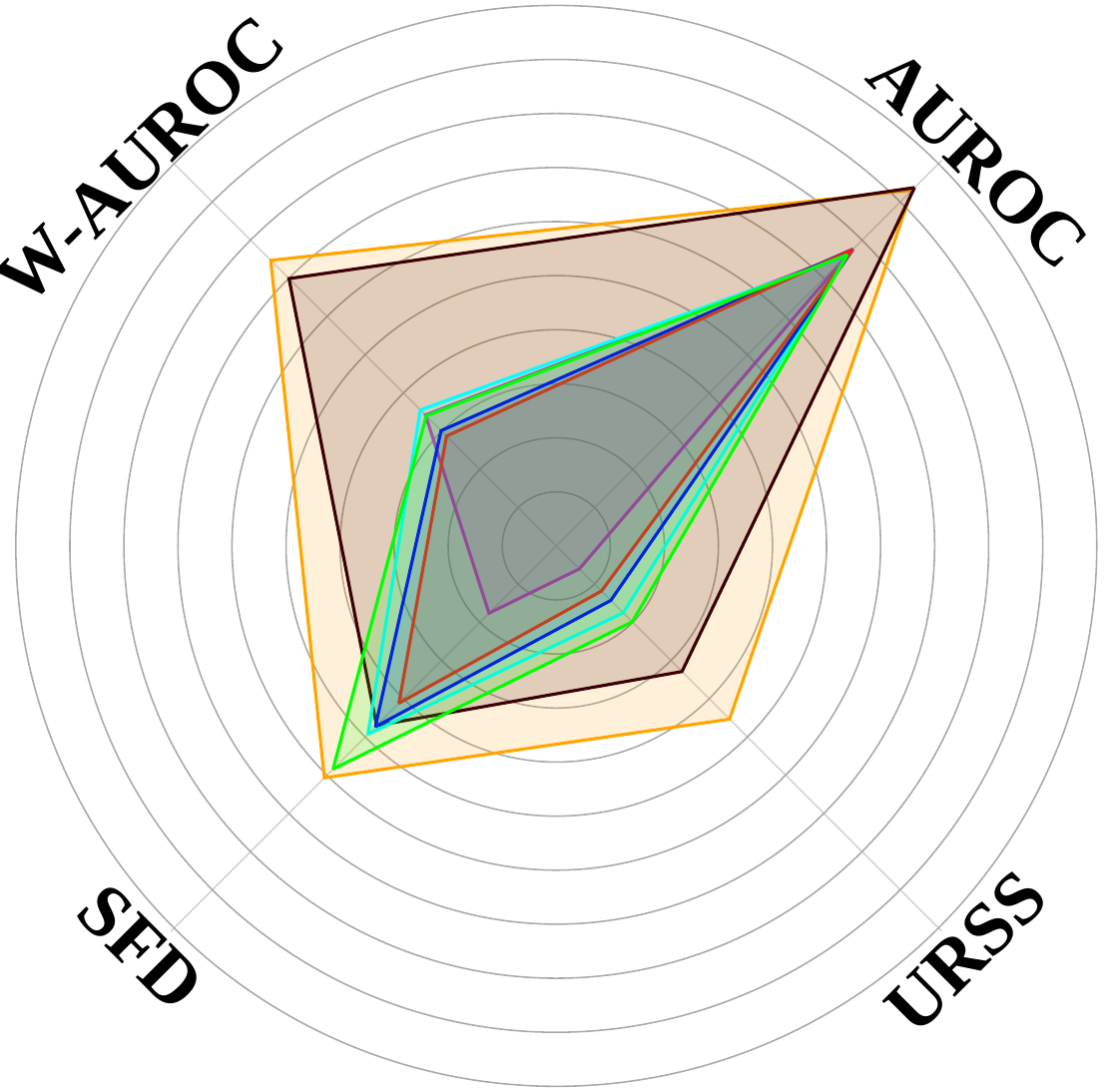} &
        \includegraphics[width=0.2\textwidth]{ leg.pdf} \\
        
    \end{tabular}

    \caption{Radar charts of comparing detectors in different generative models across all writing styles.}
    \label{fig:radarchartllms}
\end{figure*}

Figure \ref{fig:radarchartllms} shows the radar charts of comparing detectors for each LLM across all writing styles.
Table \ref{tab:smallerllms} shows the impact of different strategies on detector performance across all writing styles for smaller models from each family evaluated in this study.

\FloatBarrier
\begin{table*}[th]
\centering
\caption{Performance of detectors under paraphrasing (baseline), RMM, AWS, and RHL strategies.}
\begin{adjustbox}{max width=\textwidth}
\begin{tabular}{l|cccc|cccc|cccc}
\toprule
\textbf{LLM $\rightarrow$} & \multicolumn{4}{c|}{\textbf{Gemma2 2b}} & \multicolumn{4}{c|}{\textbf{Llama3.2 1b}} & \multicolumn{4}{c}{\textbf{Llama3.2 3b}}  \\
\textbf{Detector $\downarrow$ Metric ($\mathbf{\%}$) $\rightarrow$} & AUC & W-A & SFD & URSS & AUC & W-A & SFD & URSS & AUC & W-A & SFD & URSS  \\
\midrule
\multicolumn{13}{c}{\cellcolor{yellow!15}\textbf{Paraphrase (baseline)}} \\
\midrule
Binoculars      &97.3& 84.0& 57.6& 48.4&85.8& 33.1& 47.3& 15.7&93.6& 63.4& 64.0& 40.6\\
Fast-DetectGPT  &96.8& 87.6& 60.6& 53.0&86.4& 64.7& 58.2& 37.7&93.3& 75.4& 56.2& 42.4\\
Log-Likelihood &86.6& 40.1& 45.2& 18.1&75.9& 33.9& 51.8& 17.6&78.5& 30.8& 42.4& 13.0\\
Log-Rank        &88.1& 46.1& 63.0& 29.0&76.7& 36.9& 58.9& 21.7&79.1& 33.3& 41.3& 13.7\\
LRR             &89.5& 61.7& 54.3& 33.5&77.0& 42.9& 58.2& 24.9&79.1& 41.0& 54.7& 22.5\\
Radar            & 83.1& 49.6& 16.1& 8.0&74.2& 36.6& 14.7& 5.4&78.7& 38.9& 17.3& 6.7 \\
Rank            &79.4& 42.7& 58.8& 25.1&69.1& 29.2& 62.8& 18.3&73.6& 32.7& 56.8& 18.6\\
\midrule
\multicolumn{13}{c}{\cellcolor{yellow!15}\textbf{Random meaning-preserving mutation (RMM)}} \\
\midrule
Binoculars      &89.6& 58.6& 40.3& 23.6&80.2& 29.2& 50.9& 14.8&82.8& 39.8& 50.1& 19.9\\
Fast-DetectGPT  &88.9& 63.2& 50.8& 32.1&80.1& 46.9& 57.8& 27.1&81.8& 45.8& 49.2& 22.5 \\
Log-Likelihood &48.0& 10.5& 8.5& 0.9&42.5& 11.7& 11.8& 1.4&38.2& 6.8& 6.0& 0.4\\
Log-Rank        &52.2& 12.9& 9.3& 1.2&45.7& 13.3& 18.4& 2.5&41.0& 7.8& 7.6& 0.6\\
LRR             &67.6& 25.5& 14.5& 3.7&59.6& 20.0& 13.2& 2.6&55.3& 13.8& 10.7& 1.5\\
Radar            &91.7& 63.1& 49.3& 31.1&86.2& 50.9& 28.9& 14.7&88.8& 53.1& 30.9& 16.4\\
Rank            &55.7& 11.5& 13.2& 1.5&50.2& 8.5& 9.6& 0.8&48.7& 7.4& 9.7& 0.7\\
\midrule
\multicolumn{13}{c}{\cellcolor{yellow!10}\textbf{AI-flagged word swap (AWS)}} \\
\midrule
Binoculars      &88.4& 59.5& 49.7& 29.6&76.0& 28.1& 32.2& 9.1&81.7& 43.5& 36.5& 15.9\\
Fast-DetectGPT  &89.0& 62.3& 45.8& 28.5&78.2& 43.2& 49.7& 21.4&83.1& 47.9& 43.0& 20.6\\
Log-Likelihood &29.5& 5.9& 6.0& 0.3&25.7& 7.0& 13.1& 0.9&22.2& 3.4& 4.8& 0.2\\
Log-Rank        &31.9& 6.6& 5.5& 0.4&26.9& 7.3& 16.2& 1.2&23.5& 3.6& 4.2& 0.2\\
LRR             &45.1& 10.9& 7.2& 0.8&34.5& 7.9& 5.2& 0.4&33.5& 5.2& 4.1& 0.2\\
Radar            &89.0& 56.6& 35.5& 20.1&79.8& 38.1& 24.4& 9.3&83.5& 43.1& 25.7& 11.1 \\
Rank            & 50.5& 7.0& 12.5& 0.9&42.4& 4.9& 7.0& 0.3&44.6& 4.6& 10.3& 0.5 \\
\midrule
\multicolumn{13}{c}{\cellcolor{yellow!10}\textbf{Recursive humanification loop (RHL)}} \\
\midrule
Binoculars      &86.8& 51.1& 52.5& 26.8&73.2& 23.5& 51.9& 12.2&78.3& 35.1& 40.9& 14.4\\
Fast-DetectGPT  &86.8& 55.1& 50.6& 27.9&75.0& 34.3& 53.5& 18.4&78.9& 37.7& 45.2& 17.0\\
Log-Likelihood &28.3& 3.5& 4.3& 0.2&22.9& 4.6& 12.5& 0.6&20.0& 1.7& 3.3& 0.1\\
Log-Rank        &31.9& 4.4& 4.7& 0.2&25.3& 5.1& 12.5& 0.6&22.7& 2.1& 3.5& 0.1\\
LRR             &49.8& 10.1& 11.4& 1.2&38.9& 7.2& 6.8& 0.5&38.4& 5.1& 6.9& 0.3\\
Radar            & 89.0& 54.3& 41.5& 22.5&80.4& 37.5& 27.2& 10.2&83.1& 40.5& 26.3& 10.7 \\
Rank            &50.4& 6.7& 13.8& 0.9&41.4& 3.8& 6.6& 0.2&43.8& 4.0& 7.7& 0.3\\
\bottomrule
\end{tabular}
\end{adjustbox}
\label{tab:smallerllms}
\end{table*}

\end{document}